\documentclass{article} % For LaTeX2e
\usepackage{iclr2023_conference,times}

\usepackage{amsmath,amsfonts,bm}

\usepackage{wrapfig}

\def\eqref#1{equation~\ref{#1}}
\def\1{\bm{1}}

\DeclareMathAlphabet{\mathsfit}{\encodingdefault}{\sfdefault}{m}{sl}
\SetMathAlphabet{\mathsfit}{bold}{\encodingdefault}{\sfdefault}{bx}{n}

\usepackage{bm}
\usepackage{bbm}
\usepackage{makecell}
\usepackage{colortbl}
\usepackage{hyperref}
\usepackage{url}
\usepackage{multirow}
\usepackage{graphicx}
\usepackage{booktabs}
\usepackage{array}
\usepackage{xspace}
\usepackage{threeparttable}

\definecolor{Gray}{gray}{0.9}
\definecolor{Red}{RGB}{230, 57, 70}
\definecolor{Blue}{RGB}{0, 100, 148}

\newcommand{\Revise}[1]{\textcolor{black}{#1}}

\newcommand{\modelname}{ED-Pose\xspace}

\title{Explicit Box Detection Unifies End-to-End Multi-Person Pose Estimation}

\author{Jie Yang$^{1,2}$\thanks{This work was done when Jie Yang was intern at IDEA.}~,~Ailing Zeng$^{1}$\thanks{Corresponding author.}~,~Shilong Liu$^{1}$,~Feng Li$^{1}$,~Ruimao Zhang$^{2}$$^{\dag}$,~Lei Zhang$^{1}$ \\
$^1$International Digital Economy Academy (IDEA). \\
$^2$Shenzhen Research Institute of Big Data, The Chinese University of Hong Kong, Shenzhen \\
\texttt{\small{\{zengailing,liushilong,lifeng,leizhang\}@idea.edu.cn}}\\
\texttt{\small{\{jieyang5@link, zhangruimao@\}cuhk.edu.cn}}\\
}

\iclrfinalcopy % Uncomment for camera-ready version, but NOT for submission.
\begin{document}

\maketitle

% MaskRCNN的concept
% Explicit detection
% box size reshape >> feature map
% MaskPose modeling
\begin{abstract}

This paper presents a novel end-to-end framework with Explicit box Detection for multi-person Pose estimation, called ED-Pose, where it unifies the contextual learning between human-level (global) and keypoint-level (local) information.
%accelerate multi-person Pose estimation training and achieve high precision. 
%
%We first analyze why two-stage methods are hard to optimize when changing them to a one-stage pipeline. 
%
%Under the shared encoder, the different focus causes the optimization conflicts between global and local dependencies in an end-to-end trainable model.
%existing methods use different supervisions and representations for the global and local contextual learning.
%
Different from previous one-stage methods, ED-Pose re-considers this task as two explicit box detection processes with a unified representation and regression supervision.
First, we introduce a human detection decoder from encoded tokens to extract global features. It can provide a good initialization for the latter keypoint detection, making the training process converge fast.
%to detect person bounding boxes explicitly first
%
Second, to bring in contextual information near keypoints, we regard pose estimation as a keypoint box detection problem to learn both box positions and contents for each keypoint. 
A human-to-keypoint detection decoder adopts an interactive learning strategy between human and keypoint features to further enhance global and local feature aggregation. 
In general, ED-Pose is conceptually simple without post-processing and dense heatmap supervision.
%
% Comprehensive experiments verify the effectiveness and efficiency of the proposed ED-Pose method. 
It demonstrates its effectiveness and efficiency compared with both two-stage and one-stage methods.
Notably, explicit box detection boosts the pose estimation performance by 4.5 AP on COCO and 9.9 AP on CrowdPose. 
For the first time, as a fully end-to-end framework with a L1 regression loss, ED-Pose surpasses heatmap-based Top-down methods under the same backbone by 1.2 AP on COCO and achieves the state-of-the-art with 76.6 AP on CrowdPose without bells and whistles. Code is available at \url{https://github.com/IDEA-Research/ED-Pose}.
\end{abstract}

\section{Introduction}

\begin{figure}[h]
\vspace{-0.4cm}
\centering
 	{
 		\begin{minipage}[t]{1\linewidth}
 			\centering         
 			\includegraphics[width=1\linewidth]{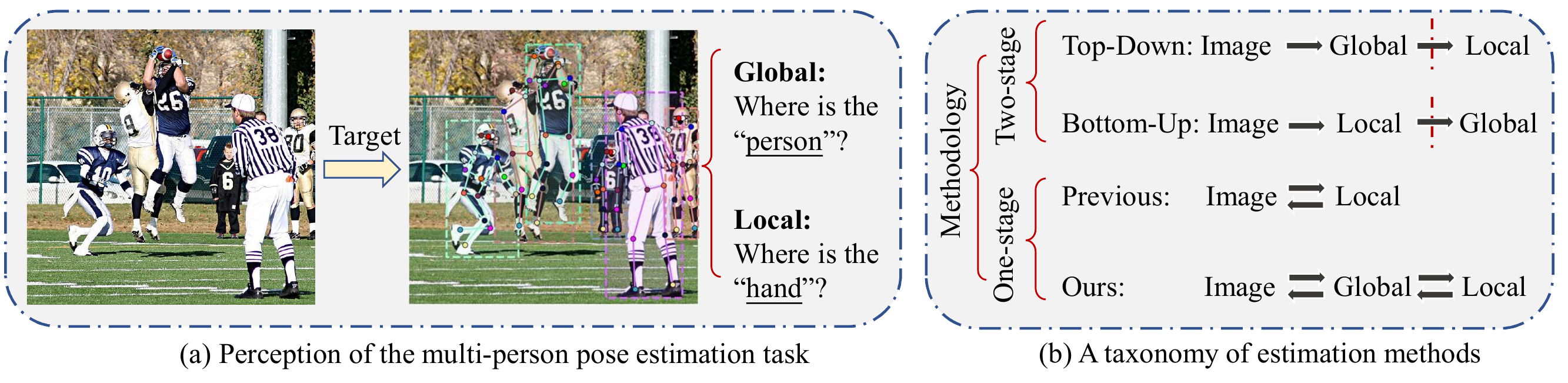}  
 		\end{minipage}
 	} 
\vspace{-0.7cm}
\caption{Illustration of (a) the perception of the pose estimation task that usually captures global and local contexts concurrently; (b) a taxonomy of existing estimators. ED-Pose (Ours) is a novel one-stage method of learning both global and local relations in an end-to-end manner.}
\label{fig:intro} 
\vspace{-0.2cm}
\end{figure}   

Multi-person human pose estimation has attracted much attention in the computer vision community for decades for its wide applications in areas of augmented reality (AR), virtual reality (VR), and human-computer interaction (HCI). Given an image, it targets to localize the 2D keypoint positions for every person in the image. 
Although many methods have been developed~\citep{xiao2018simple,sun2019deep,cheng2020higherhrnet,mao2022poseur,shi2022end}, it remains challenging and intractable for situations with heavy occlusions, hard poses, and diverse body part scales.

Intuitively, as shown in Figure~\ref{fig:intro}, this task needs to focus on both global (human-level) and local (keypoint-level) dependencies, which  concentrate on different levels of semantic granularity.
Mainstream solutions are normally two-stage methods, which divide the problem into two separate subproblems (e.g., the global person detection and local keypoint regression).
Such solutions include Top-Down (TD) methods~\citep{xiao2018simple,sun2019deep,li2021pose,mao2022poseur} which are of high performance yet with a high inference cost and Bottom-Up (BU) solutions~\citep{cao2017realtime,newell2017associative,cheng2020higherhrnet} which are fast in inference yet with relatively lower precision.
%
%They face the trade-offs between efficiency and effectiveness.
%
However, all of these methods are non-differentiable between their global and local stages due to hand-crafted operations, like Non-Maximum Suppression (NMS), Region of Interest (RoI) cropping, and keypoint grouping post-processing. 
Lately, Poseur~\citep{mao2022poseur} tries to directly apply top-down methods to an end-to-end framework and finds that there will be a significant performance drop (about \textbf{8.7} AP on COCO), indicating the optimization conflicts between the learning of global and local relations.
%, making them hard to end-to-end trainable. 

Exploring a fully end-to-end trainable method to unify the two disassembled subproblems is attractive and important.
Inspired by the success of recent end-to-end object detection methods, like DETR~\citep{carion2020end}, there is a surge of related approaches that regard human pose estimation as a direct set prediction problem. They utilize a bipartite matching for one-to-one prediction with Transformers to avoid cumbersome post-processings~\citep{li2021pose,mao2021tfpose,mao2022poseur,stoffl2021end,shi2022end}. 
%
% Most of them still follow Top-down methods to improve the second stage from the cropped global information to extract local relations. 
%
Recently, PETR~\citep{shi2022end} proposes a fully end-to-end framework to predict instance-aware poses without any post-processings and shows a favorable potential.
Nevertheless, it directly uses a pose decoder with randomly initialized pose queries to query local features from images. The only local dependency makes keypoint matching across persons ambiguous and thus leading to inferior performance, especially for occlusions, complex poses, and diverse human scales in crowded scenes.
%
%Top-down methods first detect, crop, and resize the persons and then conduct single-person pose estimation will alleviate the above issues to some extent. This is why this type of approach performs well. 
%
%to obtain the coarse keypoints and attaches a keypoint decoder behind to refine them.
%simply uses random initialized pose and keypoint queries to query the encoded features from images and then infer the keypoint positions for each person. 
%ignores the importance of detecting where are persons explicitly. 
%
Moreover, either two-stage methods or DETR-based estimators suffer from slow training convergence and need more epochs (e.g., train a model above a week) to achieve high precision. Additionally, the convergence speed of DETR-based methods is even slower than bottom-up methods~\citep{cheng2020higherhrnet}. We address the details in Sec.\ref{sec:rethink}.
%
%In fact, occlusions, complex poses, and diverse human scales in the crowd scenes are the main difficulties. 
%
%\emph{Thus, it could be significant and nontrivial to introduce explicit detection in a fully end-to-end pipeline.} 

%

% background, problem, challenges
% previous work (two-stage -> one stage)
% problems in one-stage
% our solution, and results
% contribution summary

% explicit detection in an fully end-to-end framework is nontrival.  % how to bring in explicit detection in an fully end-to-end framework
% many hand-crafted designs (NMS, ROI..) hard to end2end training.
% why two-stage?  the basic explicit detection is TD: 1. different resolutions from small persons to large persons or body parts. therefore, top-down tends to resize all detected person bounding boxes into the same large size (e.g., 384*288). but high cost and it suffers from failures when the detection fails. For BU: 2. many-to-many grouping issue. 
% why can be one-stage: anchor-free detection. 
% regard it as a set of prediction and bipart matching for the one-to-one matching. 
% hard to end2end optimize well.
% why PETR did not use detection?
% train end2end with ED to further calibrate the previous failed person detection (person AP increases)

% how to explicit obtain keypoint detection? faster and better?  

% difficulties to introduce explicit detection

Based on the above observations, this work re-considers multi-person pose estimation as two Explicit box Detection processes named ED-Pose.
We realize each box detection by using a decoder and cascade them to form an end-to-end framework, which makes the model fast in convergence, precise, and scalable. 
%
%We plug them into the decoders and combine them into an end-to-end framework, making the model fast in convergence, precise, and scalable. 
%
Specifically, to obtain global dependencies, the first process detects boxes for all persons via human box queries selected from the encoded image tokens. This simple step can provide a good initialization for the latter keypoint detection to accelerate the training convergence. 
%we disassemble the Transformer decoders into two detection decoders, a Human detection decoder, and a Human-to-Keypoint detection decoder.
%
Then, to capture local contextual relations and reduce ambiguities in the feature alignment, we regard the following pose estimation task as a keypoint box detection problem, where it learns both box positions and local contents for each keypoint.
Such an approach can leverage contextual information near a keypoint by directly regressing the keypoint box position and learning local keypoint content queries without dense supervision (e.g., heatmap). 
To further enhance the global-local interactivity among external human-human, internal human-keypoint, and internal keypoint-keypoint, we design interactive learning between human detection and keypoint detection.

Following the two Explicit box Detection processes, we can unify the global and local feature learning using the consistent regression loss and the same box representation in an end-to-end framework.
We summarize the related methods from the supervisions and representations. Compared with previous works, ED-Pose is more conceptually simple.
Notably, we find that explicit global box detection will gain \textbf{4.5} AP on COCO and \textbf{9.9} AP on CrowdPose compared with a solution without such a scheme.
In comparison to top-down methods, ED-Pose makes the human and keypoint detection share the same encoders to avoid redundant costs from human detection and further boost performance by \textbf{1.2} AP on COCO and \textbf{9.1} AP on CrowdPose under the same ResNet-50 backbone. 
%We simply use the lightweight L1 regression loss instead of heatmap supervision. 
%
Moreover, ED-Pose surpasses the previous end-to-end model PETR significantly by \textbf{2.8} AP on COCO and \textbf{5.0} AP on CrowdPose. 
% It is also faster than PETR since we simply use regression losses rather than the dense heatmap.   
% and save $70$\% training time
In crowded scenes, ED-Pose achieves the state-of-the-art with \textbf{76.6} AP (by 4.2 AP improvement over the previous SOTA~\citep{yuan2021hrformer})  without any bells and whistles (e.g., without multi-scale test and flipping).
We hope this simple attempt at explicit box detection, simplification of losses, and no post-processing to unify the whole pipeline could bring in new perspectives to further one-stage framework designs.
%We conduct comprehensive experiments to verify the effectiveness and efficiency of the proposed ED-Pose. 

\begin{table*}[h]
\caption{Comparisons of existing estimators from the losses and representations of the Human and Keypoints. Our proposed ED-Pose unifies both Human and Keypoint detection under the consistent L1 regression loss and the box representation, making the end-to-end training simple yet effective.}
\resizebox{\linewidth}{!}{
\begin{tabular}{c|c|cc|cc}
    \toprule
\multicolumn{2}{c|}{Methodology} &Human Loss& Keypoint Loss& Human Representation & Keypoint Representation \\\midrule
     \multirow{2}*{Two-Stage Methods} & Top-Down & Regression & Heatmap & $(x,y,h,w)$ & $(x,y)$\\
      & Bottom-Up & -& Heatmap &-&  $(x,y)$ \\
          \midrule
     \multirow{2}*{One-Stage Methods} & Previous & - & Regression + Heatmap & -& $(x,y)$\\
  & Ours  & Regression &Regression&  $(x,y,h,w)$ &$(x,y,h,w)$ \\
  \bottomrule
  \end{tabular}}
\label{tab:comp_loss}
\vspace{-0.3cm}
\end{table*}

\section{Related work}
\vspace{-0.2cm}
\textbf{One-stage Multi-Person Pose Estimation:} With the development of anchor-free object detectors~\citep{ZhiTian2019FCOSFC,LichaoHuang2015DenseBoxUL}, DirectPose~\citep{tian2019directpose} directly predicts instance-aware keypoints for all persons from an image. The direct end-to-end framework provides a new perspective to avoid the above cumbersome issues met in two-stage methods.
%heuristic grouping in bottom-up methods or bounding box detection and RoI operations in top-down frameworks.
%
Generally speaking, these methods densely locate a set of pose candidates, which consist of joint positions from the same person. 
%
%SPM~(\citep{nie2019single}) proposes the structured pose representation that unifies person instance and body joint position.
%
FCPose~\citep{WeianMao2021FCPoseFC} builds upon dynamic filters~\citep{XuJia2016DynamicFN} in compact keypoint heads to boost both accuracy and speed.
%eliminates the ROIs and grouping post-processing with dynamic instance-aware keypoint estimation heads. 
%
Meanwhile, Inspose~\citep{DahuShi2021InsPoseIN} designs instance-aware dynamic networks to adaptively adjust part of the network parameters for each instance.
%presents an effective one-stage solution by employing instance-aware dynamic networks. 
%
Nevertheless, these one-stage methods still need NMS to remove duplicates in the post-processing stage. 
To further remove such hand-crafted components, PETR~\citep{shi2022end} views pose estimation as a hierarchical set prediction problem and proposes the first \emph{fully} end-to-end pose estimation framework with the advent of DETR~\citep{carion2020end}.

\textbf{Detection Transformers:} For the first time, DETR~\citep{carion2020end} performs object detection in an end-to-end manner by using a set-based global loss that forces directly unique predictions via bipartite matching and a Transformer encoder-decoder architecture. 
It simplifies object detection as a direct set prediction problem, dropping multiple hand-designed components and prior knowledge.
Due to the effectiveness of DETR and its varieties (\textit{e.g.}, Deformable DETR~\citep{zhu2020deformable}), their frameworks have been widely transferred in many complex tasks, such as Mask DINO~\citep{li2022mask} for segmentation and PETR~\citep{shi2022end} for pose estimation. %redesigns the Transformer decoders to predict a set of binary masks directly for semantic segmentation.  
Following the top-down methods, PRTR~\citep{li2021pose} and TFPose~\citep{mao2021tfpose} adopt Detection Transformers to estimate the cropped single-person images as a query-based regression task. 
To capture the underlying output distribution and further improve performance in the regression paradigm, Poseur~\citep{mao2022poseur} brings Residual Log-likelihood Estimation (RLE)~\citep{li2021human} into the DETR-based top-down framework, achieving the state-of-the-art performance of regression-based methods.   
For the one-stage manner, POET~\citep{stoffl2021end} utilizes the property of DETR to directly regress the poses (instead of bounding boxes) of all person instances in an image.
Recently, PETR~\citep{shi2022end} designs a \textit{fully} end-to-end paradigm with hierarchical attention decoders to capture both relations between poses and kinematic joints.

All the aforementioned methods take advantage of Detection Transformers and densely regress a set of poses (only local relations). However, they ignore the importance of introducing explicit box detection in pose estimation to model both global and local dependencies well.

%This work tries to address the effectiveness and efficiency of the one-stage pipeline. In the decoder layers, we simply introduce explicit person bounding box detection and the interactive learning between person detection and keypoint detection to make the final pose estimation faster and better. 

\vspace{-0.3cm}
\section{Rethinking One-stage Multi-Person pose estimation}
\label{sec:rethink}
% \subsection{Problem Setup}
% \TODO{problem definition and target}
% % 介绍pose estimation问题的定义和target, under regression loss； 
\vspace{-0.2cm}
\textbf{The Necessities of One-stage Methods:}
For a long time, the two-stage paradigm dominates the mainstream methods for multi-person pose estimators. It can be generally divided into top-down methods and bottom-up methods.  
%
% \textbf{Multi-Person Pose Estimation} is often solved using either two-stage or one-stage approaches. \textbf{\textit{Two-stage methods}} can be further devided into top-down methods and bottom-up methods. 
% decouple the multi-person pose estimation into two steps:
Top-Down (TD) methods~\citep{xiao2018simple,sun2019deep,mao2022poseur} disentangle the task into detecting and cropping each person with an object detector (\textit{e.g.}, Mask RCNN~\citep{he2017maskrcnn}) in an image from the global (person-level) dependency and then conduct the single-person pose estimation via another model. 
%The former is to obtain the boxes of the person instances with an object detector. The latter is to apply the single-person estimation method to the image patch copped by boxes. 
%
They focus on local (keypoint-level) relation modeling and improving the accuracy of single-person pose estimation. However, these methods still suffer from 1). heavy reliance on the performance of the human detector, 2). redundantly computational costs for additional human detection and RoI operations, and 3). the separate training for the human detector and the corresponding pose estimator.
Instead, Bottom-Up (BU) methods~\citep{cao2017realtime,newell2017associative,cheng2020higherhrnet} first detect all the keypoints in an instance-agnostic fashion. Next, they employ a heuristical grouping algorithm to associate the detected keypoints that belong to the same person, which improves efficiency.
Even so, the complicated grouping scheme makes bottom-up methods hard to handle heavy occlusions and multiple-person scales, resulting in inferior performance.
More importantly, both of them suffer from non-differentiable optimization between the global and local features, which is not perceptive.
%
%Recent one-stage methods~\citep{tian2019directpose,WeianMao2021FCPoseFC,DahuShi2021InsPoseIN,shi2022end} aim to estimate multi-person poses directly from the input image in an end-to-end trainable framework. 
%
%Most of them still follow Top-down methods to improve the second stage from the cropped global information to extract local relations. 
%Regarding the framework, it is not only independent on the isolated pose detector used in top-down methods but also of the tricky grouping post-process used in bottom-up methods. 
Intuitively, one-stage methods could alleviate the above issues since all modules can be optimized in an end-to-end manner, and balance effectiveness and efficiency.
Interestingly, the recent DETR-based top-down method Poseur~\citep{mao2022poseur} tries to directly apply it to a one-stage framework and finds that it suffers from a significant performance drop. It might be the optimization conflicts between the global and local dependency learning using the shared encoder. Thus, how to design a one-stage framework effectively is still challenging and questionable.

\textbf{The Bottlenecks of Existing One-stage Methods:}
In terms of existing DETR-based methods, most of them still adopt the top-down framework, and improve the second single-person pose estimation via regarding it as a sequence prediction problem~\citep{WeianMao2021FCPoseFC,DahuShi2021InsPoseIN}. PETR is the first work to make the whole pipeline end-to-end without any post-processing. 
However, existing methods still have some limitations.
%does not fully tap the potential of the end-to-end pipeline.
% need explicit detection
First, all of them only utilize local dependencies to regress keypoints. Directly regressing keypoints for each person via pose queries is semantically ambiguous as the bottom-up strategies to find all keypoints from raw images instead of from cropped images.
%
%Second, they 
%Through the success of top-down methods, we learn about the importance of first capturing global information and then perform local feature extraction.   
%PETR simply uses pose and keypoint queries to query the encoded tokens from an original image, which is similar to the bottom-up strategy to find all keypoints from raw images instead of from cropped images. Localizing a person and then estimating poses in an image 
% need use previous features
Second, the pose or keypoint queries proposed in the above methods are randomly initialized without utilizing previously extracted features, making the training phase slow and ineffective.
% need keypoint box representation and global-to-local interaction
Third, the keypoint representation as a point lacks contextual information when it queries from the encoded features, leading to feature misalignment.
Last, the interactions among global-to-global, global-to-local, and local-to-local are complex, especially in crowd scenes. The current models do not pay attention to handling these complicated relations.
In this article, we try to tackle the above issues by using unified box representations and regression losses in a one-stage process.

\vspace{-0.2cm}
\section{Methodology}
\vspace{-0.3cm}
\subsection{Overview}
%
%As illustrated in Figure~\ref{fig:query}, the proposed \modelname is a fully end-to-end multi-person pose estimation framework without any post-processing.
%
%It is built on a simple architecture that contains a backbone feature extractor, Transformer encoders, Transformer decoders, and prediction heads.
%
%In general, \modelname is a fully end-to-end multi-person pose estimation framework without any post-processings. 
%
%As illustrated in Figure~\ref{fig:query}, it is built on a simple architecture that contains a backbone, Transformer encoders, Transformer decoders, and prediction heads. Transformer decoders consist of a \textit{Human Detection Decoder} and a \textit{Human-to-Keypoint Detection Decoder}, which explicitly detect the person and keypoint. They will be described in detail in Sec.\ref{sec:box-decoder} and Sec.~\ref{sec:box-kpt-decoder}, respectively.
%
%
\begin{figure}[h]	
\vspace{-0.4cm}
\centering
 	{
 		\begin{minipage}[t]{1.0\linewidth}
 			\centering         
 			\includegraphics[width=1.0\linewidth]{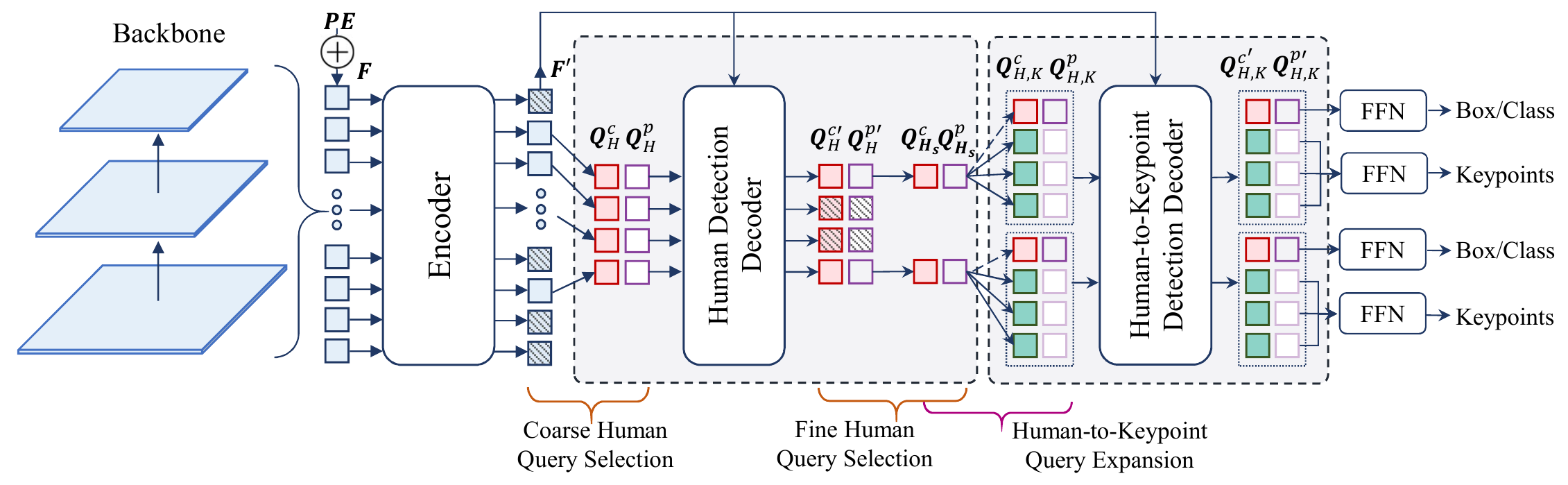}   
 		\end{minipage}
 	} 
\vspace{-0.8cm}
\caption{ The overview architecture of our ED-Pose, which contains a Human Detection Decoder and a Human-to-Keypoint Detection Decoder to detect human and keypoint boxes explicitly.
%
%The overview of the proposed fully end-to-end trainable ED-Pose framework. Transformer decoders consist of a Human Detection Decoder and a Human-to-Keypoint Detection Decoder to detect human and keypoint boxes explicitly.} 
}
\label{fig:query} 
\vspace{-0.3cm}
\end{figure}

As illustrated in Figure~\ref{fig:query}, the proposed \modelname is a fully end-to-end multi-person pose estimation framework without any post-processing.
Given an input image, we first utilize a backbone 
% (e.g., ResNet50~\citep{He_2016_resnet})  
to extract multi-scale features and obtain tokenized representations $\mathbf{F}$,
and then feed them into the Transformer encoder (e.g., deformable attention modules~\citep{zhu2020deformable}) with Positional Embeddings ($\mathbf{PE}$)
to calculate the refined tokens $\mathbf{F}'$.
%
%to merge and refine features $\mathbf{F}$ to encoded tokens $\mathbf{F}'$. 
%The Transformer encoders are the . 
%
To improve the efficiency of the following decoding process, we conduct coarse human query selection from $\mathbf{F}'$ to obtain the sparse human content queries $\mathbf{Q}_{H}^c$.
%
% Then both $\mathbf{Q}_{H}^c$ and the corresponding position queries $\mathbf{Q}_{H}^p$ are feed into the \textit{Human Detection Decoder} to update into the corresponding queries $\mathbf{Q}_{H}^{c'}$ and $\mathbf{Q}_{H}^{p'}$, respectively. 
Then we utilize $\mathbf{Q}_{H}^c$ to generate the corresponding position queries $\mathbf{Q}_{H}^p$ via a Feed-Forward Network (FFN) and feed both $\mathbf{Q}_{H}^c$ and  $\mathbf{Q}_{H}^p$ into the \textit{Human Detection Decoder} to update into the corresponding queries $\mathbf{Q}_{H}^{c'}$ and $\mathbf{Q}_{H}^{p'}$, respectively (see Sec.\ref{sec:box-decoder}). 
%the candidate human box positions.
%
%The decoder will output the updated 
%
%Please refer to Sec.\ref{sec:box-decoder} for more technique details. 
%
For each output human content query, we attach a human box regression and class entropy supervision ($L_h^l$ and $L_c^l$) at the $l$-th decoder layer.
%
%
%
%the \textit{Human Detection Decoder}, we first conduct sparse selection on $\mathbf{F}'$ to obtain the human content query $\mathbf{Q}_{H}^c$ to reduce redundant queries and improve efficiency. Then, we input $\mathbf{Q}_{H}^c$ and $\mathbf{Q}_{H}^p$ into the decoder to reason human box positions.
%%, both of which are acquired from encoded features $\mathbf{F}'$ using a sparse human query selection. 
%
%After reasoning the human box information, 
%The decoder will output the updated human content query $\mathbf{Q}_{H}^{c'}$ and positional query $\mathbf{Q}_{H}^{p'}$.
%
%% we adopt a sparse person box selection strategy from encoded features to obtain the box content query $\mathbf{Q_{box}^c}$ as well as the box positional query $\mathbf{Q_{box}^p}$ for higher efficiency and better query initialization compared with random initialization. 
%
Next, we further perform fine human query selection to discard the redundant human queries, obtaining content queries $\mathbf{Q}_{H_s}^{c}$ and position queries $\mathbf{Q}_{H_s}^{p}$.
We initialize keypoint queries based on these retained high-quality human queries and concatenate the human and keypoint queries together to form human-keypoint queries.
We name such a scheme as human-to-keypoint query expansion, and the obtained human-keypoint content queries $\mathbf{Q}_{H,K}^{c}$ and position queries $\mathbf{Q}_{H,K}^{p}$ are fed into \textit{Human-to-Keypoint Detection Decoder}.
In this way, we realize the keypoint detection at an instance level and update these queries to $\mathbf{Q}_{H,K}^{c'}$ and $\mathbf{Q}_{H,K}^{p'}$ layer by layer (see Sec.\ref{sec:box-kpt-decoder}).
Also, the keypoint and human box regression loss and class loss ($L_k^l$, $L_h^l$ and $L_c^l$) are added to the $l$-th decoder layer.
%
%Please refer to Sec.\ref{sec:box-kpt-decoder} for detailed implementation.
%
Finally, we use several FFN heads to regress the keypoint positions in each human box.

\textbf{Loss:}
Following the DETR~\citep{carion2020end}, we employ a set-based Hungarian loss that forces a unique prediction for each ground-truth box and keypoint. Our overall loss functions contain classification $L_c$, human box regression $L_h$, and keypoint regression loss $L_k$. Notably, $L_k$ simply consists of the normal L1 loss and the constrained L1 loss named Object Keypoint Similarity (OKS) loss~\citep{shi2022end} without any dense supervision (e.g., heatmap).

\begin{figure}[ht]	
\centering
\vspace{-0.3cm}
    \centering
    \includegraphics[width=1\textwidth]{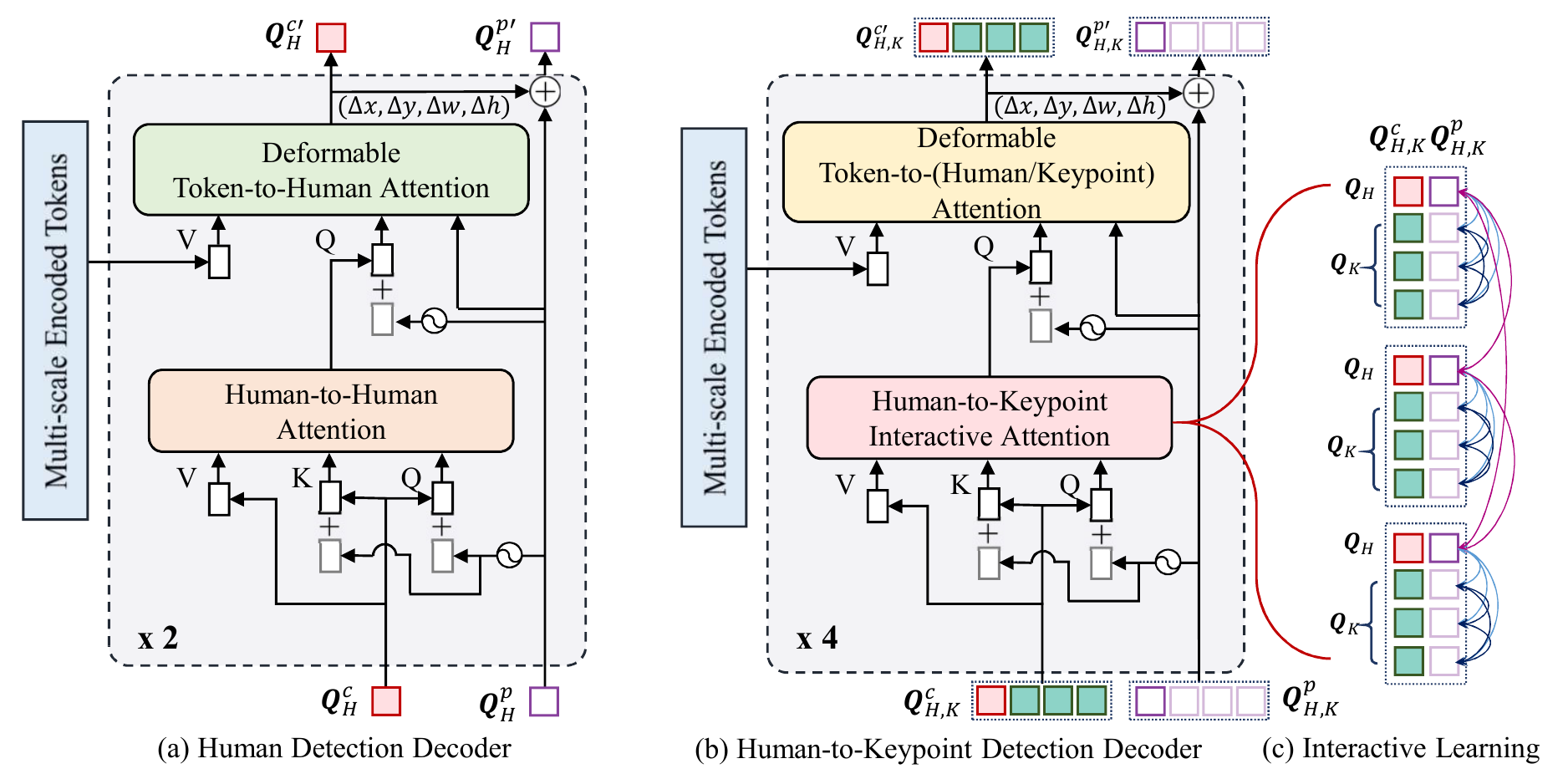}
    \vspace{-0.7cm}
    \caption{The detailed illustration of (a) Human Detection Decoder, (b) Human-to-Keypoint Detection Decoder and (c) the detailed Interactive Learning in Human-to-Keypoint Detection Decoder. }
    \label{fig:decoder}
    \vspace{-0.3cm}
\end{figure}

\subsection{Human Detection Decoder}
\label{sec:box-decoder}
\vspace{-0.2cm}
In Figure~\ref{fig:decoder} (a), the proposed Human Detection Decoder aims to predict the candidate bounding box positions for each person in the image, as well as the corresponding content representation by leveraging image context from all encoded tokens.
%
%reason the bounding box positions for each person in the image and the corresponding content information under the whole image context (\textit{i.e.}, encoded features).  
%
%
% Similar to DAB-DETR \citep{liu2022dab}, we formulate the position queries into 4D anchor points $(x, y, w, h)$.
%
Given $N$ input human content queries $\mathbf{Q}_{H}^c\in \mathbb{R}^{N\times D}$ and human position queries $\mathbf{Q}_{H}^p \in \mathbb{R}^{N\times 4}$, where $D$ is the channel dimension, the Human Detection Decoder outputs $N$ refined human box positions $\mathbf{Q}_{H}^{p'}$ and content representations $\mathbf{Q}_{H}^{c'}$.
%by updating these queries.
% $\mathbf{Q}_{H}^p$ and $\mathbf{Q}_{H}^c$ to the same shape $\mathbf{Q}_{H}^{p'}$ and $\mathbf{Q}_{H}^{c'}$ through \emph{Human-to-Human Attention} layers and \emph{Deformable Feature-to-Human Attention} layers, respectively.

% \setlength{\tabcolsep}{2pt}
% \begin{wrapfigure}{r}{1\textwidth}

\textbf{Coarse-to-Fine Human Query Selection}: 
\label{sec:N-to-M}
Unlike previous work~\citep{shi2022end} to randomly initialize the input queries,
we adopt the query selection (QS) strategy for better initialization~\citep{zhang2022dino}. 
Specifically, we propose a coarse-to-fine scheme to progressively select high-quality human queries.
%
%
%Specially, we propose a coarse-to-fine human QS strategy including \textit{coarse human QS} and \textit{fine human QS}, as shown in Figure~\ref{fig:query}. 
%
%
In practice, the \textit{coarse human QS} selects first from a large number of refined tokens $\mathbf{F}' \in \mathbb{R}^{T \times D}$,
%
%as the image priors to enhance the input information and avoid redundancy for the Human Detection Decoder, 
%
where $T$ is the number of the tokens (e.g., about $15$K).
Thus, the human content queries $\mathbf{Q}_{H}^c\in \mathbb{R}^{N\times D}$ are initialized by the top-$N$ refined tokens' features, ranked by their instance classification scores. 
%
%
%Accordingly, the human content query $\mathbf{Q}_{H}^c\in \mathbb{R}^{N\times D}$ is directly initialized by the top-$N$ encoded tokens' feature by ranking the instance-level classification scores. 
%
%
Here $N$ indicates the number of candidate human boxes (e.g., $900$). %
Then we use these selected human content queries to calculate their position queries $\mathbf{Q}_{H}^p\in \mathbb{R}^{N\times 4}$ by employing a simple detection head.
%
%Meanwhile, the human positional query is initialized by the corresponding coarse human box coordinates, which are obtained by employing a detection head upon the $N$ selected encoded tokens. 
%
%
After passing through the Human Detection Decoder, we further conduct the \textit{fine human QS} by only retaining $M$ (e.g., $100$) refined human content queries and their position queries according to the classification scores, denoted as $\mathbf{Q}_{H_s}^{c}\in \mathbb{R}^{M\times D}$ and $\mathbf{Q}_{H_s}^{p}\in \mathbb{R}^{M\times 4}$.
%
%
%After passing through the Human Detection Decoder, we further conduct the \textit{fine human QS} by only retaining $M$ (e.g., $100$) enhanced human content queries in $\mathbf{Q}_{H}^{c'}$ according to their classification scores, 
%
%and obtain the $\mathbf{Q}_{H_s}^{c}\in \mathbb{R}^{M\times D}$ and $\mathbf{Q}_{H_s}^{p}\in \mathbb{R}^{M\times 4}$ respectively as the input of the Human-to-Keypoint Decoder. 

%employ the \textit{fine human QS} via the top-$M$ (e.g., $100$) classification scores to reduce the human queries into $\mathbf{Q}_{H_s}^{c}\in \mathbb{R}^{M\times D}$ and $\mathbf{Q}_{H_s}^{p}\in \mathbb{R}^{M\times 4}$ for Human-to-Keypoint Decoder.

\textbf{Human Box Detection}:
As illustrated in Figure~\ref{fig:decoder} (a),
we combine $\mathbf{Q}_{H}^c$ with $\mathbf{Q}_{H}^p$ and feed them into Human-to-Human Attention (\textit{i.e.} a self-attention layer) to calculate the relations among human queries, and output contextual enhanced ones.
%
%
%As shown on the left of figure~\ref{fig:decoder}, the human content queries $\mathbf{Q}_{H}^c$ added with the human box positional embedding $\mathbf{Q}_{H}^p$ into a self-attention module to learn the correlations among humans (called Human-to-Human Attention) and enhance the instance-level representation capability for the human content query.  
%
Motivated by the deformable DETR~\cite{zhu2020deformable}, these enhanced human content queries, together with their position queries are further fed into Deformable Token-to-Human Attention (\textit{i.e.} a cross-attention layer) to update human queries by conducting interaction with multi-scale tokens.
Next, to adjust the human box positions, we leverage updated human queries to calculate the 4D offsets following the DAB-DETR~\citep{liu2022dab}, and add them back to previous human position queries. 
In this way, the Human Detection Decoder refine the human content and position queries progressively by stacking multiple aforementioned layers and output $\mathbf{Q}_{H}^{c'}\in \mathbb{R}^{N\times D}$ and $\mathbf{Q}_{H}^{p'}\in \mathbb{R}^{N\times 4}$.

\subsection{Human-to-Keypoint Detection Decoder}
\label{sec:box-kpt-decoder}
\vspace{-0.2cm}
To unify both human and keypoint representations as the box and facilitate follow-up interactive learning, we regard multi-person pose estimation as the multiple set keypoint box detection problems.
Based on the selected high quality human queries $\mathbf{Q}_{H_s}^{p}$ and  $\mathbf{Q}_{H_s}^{c}$, 
we initialize multiple sets of keypoint queries (\textit{i.e.} including both content and position queries) through the Human-to-Keypoint Query Expansion process,
and concatenate the human and keypoint queries together as inputs of the Human-to-Keypoint Detection Decoder (Figure.~\ref{fig:decoder} (b)) to calculate the precise keypoint positions for each person.

\textbf{Human-to-Keypoint Query Expansion}: 
After obtaining $\mathbf{Q}_{H_s}^{p}$ and  $\mathbf{Q}_{H_s}^{c}$, the keypoint content query and positional query are initialized by the relevant human query instead of random initialization.  
In practice, we first initialize a set of learnable keypoint embeddings $\mathbf{V_e} \in \mathbb{R}^{1 \times K\times D}$ where $K$ is the total number of keypoints (e.g., 17).
To concretize these embeddings to multiple specific persons' keypoint queries, we broadcast the first dimension of $\mathbf{V_e}$ to the number of human queries $M$ and then add it with $\mathbf{Q}_{H_s}^{c} \in  \mathbb{R}^{M\times 1 \times D}$ to obtain $M$ set of keypoint content queries.
For a specific set of keypoint position queries,
we separate the initialization process into center coordinate initialization (e.g., \{x, y\}) and box size initialization (e.g., \{w, h\}).
For the former, we adopt an FFN to regress all of the $K$ positions' coordinates by employing the corresponding human content query.
For the latter, the sizes of $K$ boxes are conditioned by the width and height of the corresponding human box via dot-multiplying dynamic weights $\textbf{W} \in \mathbb{R}^{K\times 2}$.
After that, we consider each human box and the corresponding set of keypoint boxes as a whole and generate the human-keypoint content queries $\mathbf{Q}_{H,K}^{c} \in  \mathbb{R}^{(M+M*K)\times D}$ and position queries $\mathbf{Q}_{H,K}^{p} \in  \mathbb{R}^{(M+M*K)\times 4}$ for further predictions.

\textbf{The Interactive Learning between Human and Keypoint Detection}: 
As illustrated in Figure~\ref{fig:decoder} (b) and (c), 
after generating human-keypoint queries, we feed them into Human-to-Keypoint Interactive Attention to learn relations from internal human-keypoint, internal keypoint-keypoint, and external human-human. 
%
%extensively.
%
%after preparations of input queries, we design a Human-to-Keypoint Interactive Attention scheme to learn relations from internal human-keypoint, internal keypoint-keypoint, and external human-human.
%
Such an interactive learning method between global and local feature aggregations has been ingeniously introduced to ensure the effectiveness of keypoint feature extraction. 
In practice, we find that the fully connected interactive learning with external keypoint-keypoint would cause a large disturbance, especially in crowded situations, similar to problems encountered in bottom-up methods. 
In contrast, our external human-human interactions effectively propagate global context across different candidates and distinguish their relations clearly.
Then we take the enhanced human-to-keypoint content queries to conduct the interaction with multi-scale encoded tokens to obtain updated human-to-keypoint queries. 
By stacking multiple layers illustated in Figure~\ref{fig:decoder} (b), we finally obtain the refined human-to-keypoint queries, denoted as $\mathbf{Q}_{H, K}^{c'}$ and  $\mathbf{Q}_{H, K}^{p'}$.

\section{Experiments}
\vspace{-0.2cm}
Due to the page limit, we leave the detailed experiment setup 
%
%(dataset descriptions and implementation details)
%
in \ref{sec:app_exp}, comparison results on COCO test-dev in \ref{sec:app_coco}, comparisons on human detection in \ref{sec:app_det}, more qualitative results and analyses in \ref{sec:app_viz}, and more discussion for ED-Pose (in Sec.~\ref{sec:app_dis}).

\vspace{-0.1cm}
\subsection{Results on CrowdPose}
\vspace{-0.2cm}
We first verify the effectiveness of ED-Pose with other state-of-the-art methods on the CrowdPose \texttt{test} set in Table ~\ref{tab:crowdpose}.
Compared with the top-down methods, we surpass the SimpleBaseline~\citep{xiao2018simple} under the same backbone by $9.1$ AP.  we also show superiority on all previous methods combined with Swin-L~\citep{liu2021Swin} backbone and outperform PETR by $1.5$ AP under the same backbone. When we enlarge the scales of backbones, ED-Pose will achieve the state-of-the-art 76.6 AP without multi-scale and flip tests.
%The results of our ED-Pose and other state-of-the-art methods on the CrowdPose \texttt{test} set are shown in Table ~\ref{tab:crowdpose}.
%Unlike top-down methods that have lost their superiority in crowded scenes, our approach shows its robustness and achieves a $73.1$ AP score using Swin-L~\citep{liu2021Swin} as a backbone, which surpasses the latest bottom-up method SWAHR~\citep{luo2021rethinking} and one-stage method PETR~\citep{shi2022end}. 
%Moreover, we achieve
%a new state-of-the-art $76.6$ AP on CrowdPose without multi-scale and flip tests.

\begin{table*}[h]
\vspace{-0.2cm}
  \centering
  \caption{Comparisons with state-of-the-art methods on CrowdPose \texttt{test} dataset. \textbf{TD, BU, OS} mean top-down, bottom-up, and one-stage methods, respectively. We use ``HM.'' and ``R.'' for heatmap-based losses and regression losses. $\dag$ denotes the flipping test. The model with * is pre-trained on Objects365~\citep{ShuaiShao2019Objects365AL} with $5$ feature scales. The \underline{underlined} highlights the compared results. The \textbf{best results} are highlighted in {\textbf{bold}}.}
  \resizebox{0.8\linewidth}{!}{
  \begin{tabular}{ll|l|ccc|ccc}
    \toprule
       &{\textbf{Method}}  & Loss & ${\rm AP}$ & ${\rm AP}_{50}$ & ${\rm AP}_{75}$ & ${\rm AP}_{E}$ & ${\rm AP}_{M}$ & ${\rm AP}_{H}$\\
       \midrule
       %\multicolumn{8}{c}{\textbf{Top-down Methods}} \\
    % \midrule
	\parbox{2.5mm}{\multirow{4}{*}{\rotatebox[origin=c]{90}{\textbf{TD}}}}
    % AlphaPose & - & 61.0 & 81.3 & 66.0 & 71.2 & 61.4 & 51.1  \\
    &Sim.Base. (ResNet-50)  & HM. & \underline{60.8} & 81.4 & 65.7 & 71.4 &61.2 & 51.2  \\
    &HRNet (HRNet-w48) $\dag$ & HM. & 71.3 & 91.1 & 77.5 & 80.5 & 71.4 & 62.5\\
    &TransPose-H & HM. & 71.8 & 91.5 & 77.8 & 79.5 & 72.9 & 62.2 \\
    &HRFormer-B & HM. & 72.4 & 91.5 & 77.9 & 80.0 & 73.5& 62.4\\
    \midrule
    \parbox{2.5mm}{\multirow{3}{*}{\rotatebox[origin=c]{90}{\textbf{BU}}}}
    %\multicolumn{8}{c}{\textbf{ Methods}} \\
    %\midrule
    &HrHRNet-w32$\dag$  & HM.  &  65.9  & 86.4 & 70.6 & 73.3 & 66.5 & 57.9\\
    &DEKR (HrHRNet-w32)$\dag$& HM.  & 65.7 & 85.7 & 70.4 & 73.0 & 66.4 & 57.5 \\
    % PINet \TODO{backbone}$\dag$ & HM. & 68.9 &  88.7 & 74.7 & 75.4 & 69.6 & 61.5\\
    &SWAHR (HrHRNet-w32)$\dag$& HM. & 71.6 & 88.5 & 77.6 & 78.9 & 72.4 & 63.0\\
    \midrule
    \parbox{2.5mm}{\multirow{4}{*}{\rotatebox[origin=c]{90}{\textbf{OS}}}}
    % \multicolumn{8}{c}{\textbf{One-stage Methods}} \\
    % \midrule
    &PETR (Swin-L)  & R.+HM.  & \underline{71.6} & 90.4 & 78.3 & 77.3 & 72.0 & 65.8 \\
       \cmidrule{2-9}
    &ED-Pose (ResNet-50)  & R.  &  69.9{\color{Red}$\uparrow_{9.1}$} & 88.6& 75.8& 77.7& 70.6&60.9 \\
    &ED-Pose (Swin-L)  & R.  &  73.1{\color{Red}$\uparrow_{1.5}$} & 90.5& 79.8& 80.5& 73.8&63.8\ \\
    &ED-Pose (Swin-L*) & R.  &  \textbf{76.6}{\color{Red}$\uparrow_{5.0}$}& \textbf{92.4}& \textbf{83.3}& \textbf{83.0}& \textbf{77.3}& \textbf{68.3} \ \\
  \bottomrule
\end{tabular}}
\label{tab:crowdpose}
\vspace{-0.4cm}
\end{table*}

\begin{table*}[h]
\centering
\caption{Comparisons with state-of-the-art methods on COCO \texttt{val2017} dataset. $\dag$ denotes the flipping test. $\ddag$ removes the prediction uncertainty estimation in Poseur as a fair regression comparison. The \underline{underlined} highlights the compared results. 
% \emph{Emphasized bold number} in inference time is tested by us, otherwise, from PETR's paper directly or the MMPose official results.
\Revise{The inference time of all methods is tested on an A100, except that the detector of top-down methods is tested by the MMdetection (\textit{i.e.}, $45$ms).}
% The ``RLE'' is a learnable regression-based loss.  
%The \textbf{best results} are highlighted in {\textbf{bold}}.
}
\resizebox{0.9\linewidth}{!}{
  \begin{tabular}{p{0.4cm}l|l|l|l|ccc|cc|c}
    \toprule
       &\multicolumn{2}{c|}{\textbf{Method}} & Backbone & Loss & ${\rm AP}$ & ${\rm AP}_{50}$ & ${\rm AP}_{75}$ & ${\rm AP}_{M}$ & ${\rm AP}_{L}$ & Time [ms]\\
    \midrule
    \parbox{2.5mm}{\multirow{8}{*}{\rotatebox[origin=c]{90}{\textbf{Two-stage}}}}
    % \multicolumn{10}{c}{\textbf{Two-stage Methods}} \\
    % \midrule
    &\multirow{6}*{{\rotatebox{90}{\textbf{TD}}}}
    & Sim.Base.$^{\dag}$ & ResNet-50 & HM. &70.4 &88.6 & 78.3 & 67.1 & 77.2 & 45$+${{86}}\\
    && HRNet$^{\dag}$ &  HRNet-w32 & HM. & 74.4 & 90.5& 81.9& 70.8& 81.0& 45$+$\Revise{{112}} \\
    &&PRTR$^{\dag}$ & ResNet-50 & R. & 68.2 & 88.2 & 75.2 & 63.2 & 76.2 & 45$+$\Revise{{85}}\\
    &&Poseur &  ResNet-50 & RLE$^{\ddag}$ & 70.0 & - & -&- &- &  45$+${82} \\ 
    &&Poseur &  ResNet-50 & RLE & 74.2 & 89.8 & 81.3&\textbf{71.1} &80.1 &  45$+${{82}} \\ 
    \cmidrule{2-11}
    &\multirow{3}*{{\rotatebox{90}{\textbf{BU}}}}    
    & HrHRNet$^{\dag}$ &  HRNet-w32 & HM. & 67.1 & 86.2 &73.0 &61.5 &76.1 &  {{322}}\\
    && DEKR$^{\dag}$  &  HRNet-w32 & HM. & 68.0 & 86.7 & 74.5 & 62.1 & 77.7  & -\\
    && SWAHR$^{\dag}$ &  HRNet-w32 & HM. & 68.9 & 87.8 & 74.9 & 63.0 & 77.4 & - \\
        \midrule
    \parbox{2.5mm}{\multirow{4}{*}{\rotatebox[origin=c]{90}{\textbf{E2E TD}}}}
    %\multicolumn{10}{c}{\textbf{End-to-End Top-down Methods}} \\     
    && Mask R-CNN &  ResNet-101 & HM. &66.0 & 86.9 & 71.5 &- &- & -\\
    &&PRTR$^{\dag}$ & ResNet-101 & HM. & 64.8 &85.1 &70.2& 60.4 &73.8& -\\
    &&Poseur$^{\dag}$ & ResNet-101 & RLE  & 68.6 & 87.5 & 74.8 & - & - & - \\
    &&Poseur$^{\dag}$ & HRNet-w48 & RLE  & 70.1&88.0 & 76.5 & - &- & -\\
    \midrule
    \parbox{2.5mm}{\multirow{6}{*}{\rotatebox[origin=c]{90}{\textbf{OS}}}}
    % \multicolumn{10}{c}{\textbf{One-stage Methods}} \\
    % \midrule
    &&InsPose  & ResNet-50 &  R.+HM.  & 65.2& 87.2 & 71.3 & 60.6 & 72.2 & \Revise{78}\\
    &&PETR & ResNet-50 & R.+HM. & \underline{68.8}	& 87.5	& 76.3	& 62.7	& 77.7 & {105}  \\
    % &PETR & ResNet-101 &R.+HM. & 70.0 & 88.5	& 77.5 & 63.6 & 79.4 & \underline{123}  \\
    &&PETR & Swin-L & R.+HM.  & \underline{73.1}	& 90.7 & 80.9 & 67.2 & 81.7& {206} \\
     \cmidrule{3-11}
    &&ED-Pose & ResNet-50 & R.  &  71.6{\color{Red}$\uparrow_{2.8}$}&  89.6 & 78.1 & 65.9 & 79.8 & {\textbf{51}}{\color{Red}$\downarrow_{51.4\%}$}\\
    % &ED-Pose & ResNet-101 & R.  &   & & & & & \\
    %         % &ED-Pose & ResNet-152 & Reg.  &   & & & & & \\
    &&ED-Pose & Swin-L & R.  &  74.3{\color{Red}$\uparrow_{1.2}$} & 91.5 & 81.6 & 68.6 & 82.6 & {{88}}{\color{Red}$\downarrow_{57.3\%}$}  \\
    &&ED-Pose & Swin-L* & R.  &  \textbf{75.8}{\color{Red}$\uparrow_{2.7}$} & \textbf{92.3} & {82.9} & 70.4& \textbf{83.5}& {{142}}{\color{Red}$\downarrow_{31.1\%}$} \\
  \bottomrule
\end{tabular}}
\label{tab:sota}
\vspace{-0.4cm}
\end{table*}

\subsection{Results on COCO}
\vspace{-0.2cm}
We further make comparisons with the state-of-the-art
methods on COCO \texttt{val2017} and \texttt{test-dev} in Table.~\ref{tab:sota} and Table.~\ref{tab:one-stage} respectively. 
In general, our ED-Pose outperforms all existing bottom-up methods and one-stage methods under the same backbone without any tricks, even the dense heatmap-based top-down methods. The proposed method achieves $71.6$ AP (by $2.8$ AP improvement) with a $51.4\%$ inference time reduction via the ResNet-50 backbone. 
Additionally, our best model with the Swin-L* backbone achieves a $75.8$ AP, showing a consistent performance improvement. The improvements in the \texttt{test-dev} are similar.

\vspace{-0.2cm}
\subsubsection{Comparison of Effectiveness}
\vspace{-0.2cm}
\textbf{Comparison with one-stage methods}: Our method significantly outperforms all existing one-stage methods, especially in Table.~\ref{tab:one-stage}, such as DirecetPose~\citep{tian2019directpose}, FCPose~\citep{WeianMao2021FCPoseFC}, Inspose~\citep{DahuShi2021InsPoseIN}, CenterNet~\citep{zhou2019objects}, and PETR~\citep{shi2022end}, showing that ED-Pose with explicit box detection is an effective solution for the end-to-end framework.
%Our proposed ED-Pose has $2.2$ AP gains compared with the recent fully end-to-end method PETR~\citep{shi2022end} with a ResNet-50 backbone.
% Compared with the recent fully end-to-end method PETR~\citep{shi2022end}, the performance of our ED-Pose is $2.2$ points higher with ResNet-50 on COCO \texttt{test-dev} set. 
% \vspace{-0.1cm}

\textbf{Comparison with two-stage methods}: For bottom-up methods, ED-Pose outperforms state-of-the-art methods by a large margin, such as HigherHRNet~\citep{cheng2020higherhrnet}, DEKR~\citep{geng2021bottom}, and SWAHR~\citep{luo2021rethinking}. Specifically, ED-Pose significantly surpasses the recently proposed SWAHR by $\textbf{3.7}$ AP ($71.6$ AP \textit{vs}. $67.9$ AP) even with a much smaller backbone (ResNet-50 \textit{vs}. HRNet-w32).
For top-down methods, ED-Pose is $1.2$ AP and $3.4$ AP higher than SimpleBaseline~\citep{xiao2018simple} (Sim.Base.) and PRTR~\citep{li2021pose}, respectively. To the best of our knowledge, this is the first time that a fully end-to-end framework can surpass top-down methods. 
% However, ED-Pose is still lower than poseur's result Since this method adopt a tricky regression loss-RLE~\citep{li2021human}.

\textbf{Comparison with end-to-end top-down methods}:
Poseur~\citep{mao2022poseur} extends its framework to end-to-end human pose estimation with Mask-RCNN. Similarly, PRTR presents its end-to-end variant. However, compared to their two-stage paradigms, their end-to-end frameworks produce substantially inferior performance. Our fully end-to-end framework achieves much higher AP scores than theirs, indicating the effectiveness of our proposed methods.

\begin{figure}[h]	
\vspace{-0.2cm}
\centering
 	{
 		\begin{minipage}[t]{1\linewidth}
 			\centering         
 			\includegraphics[width=1\linewidth]{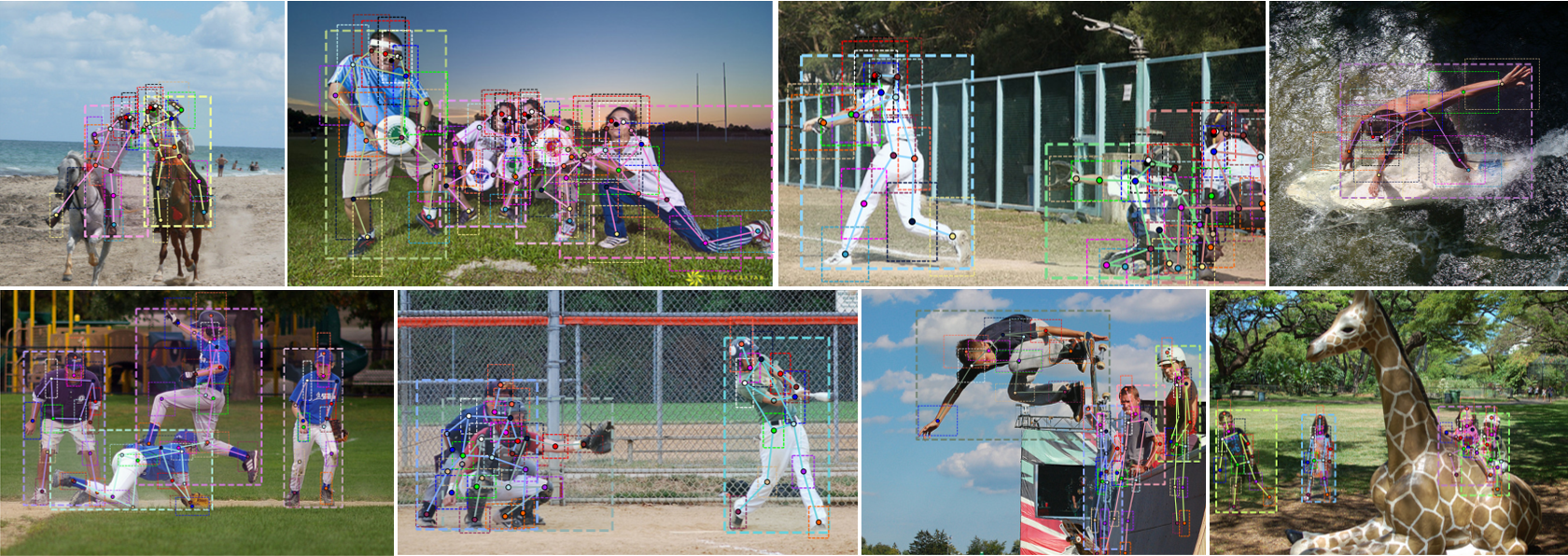}  
 		\end{minipage}
 	} 
\vspace{-0.6cm}
\caption{Qualitative results of ED-Pose on COCO (the first row) and CrowdPose (the second row). We present both explicitly detected person boxes and keypoint boxes to understand how they work.}
\label{fig:viz1} 
% \vspace{-0.1cm}
\end{figure}

\textbf{Qualitative results:} With explicit box detection, ED-Pose can perform well on a wide range of poses, containing viewpoint change, occlusion, motion blur, and crowded scenes. Fig.~\ref{fig:viz1} demonstrates the results of the detected person and the corresponding keypoints. As can be observed, the human boxes are precise under many severe cases, thus they can provide effective global human information for further keypoint detection. The local regions of the keypoint boxes are also reasonable for bringing abundant contextual information near the keypoint.

\subsubsection{Comparison of Efficiency}
\begin{figure}[h]	
\vspace{-0.4cm}
\centering
 	{
 		\begin{minipage}[t]{0.9\linewidth}
 			\centering         
 			\includegraphics[width=1\linewidth]{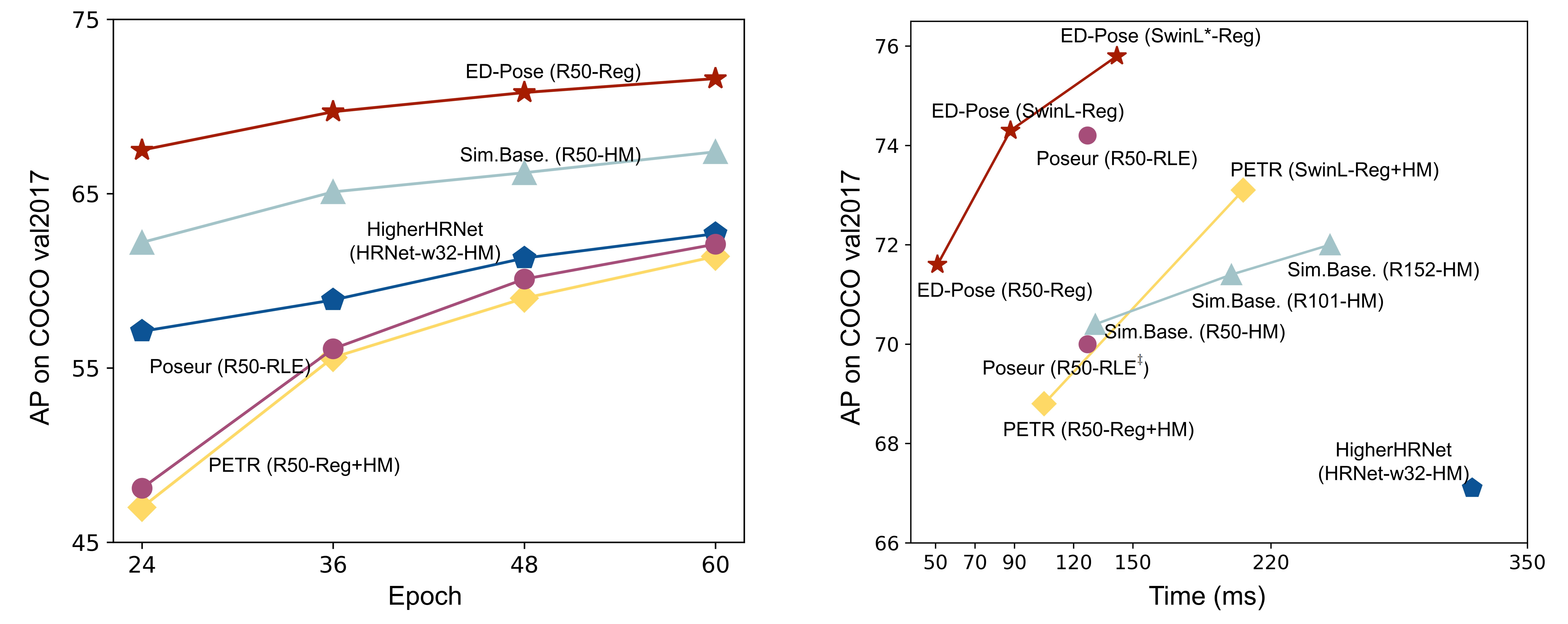}  
 		\end{minipage}
 	} 
\vspace{-0.4cm}
\caption{Comparisons of convergence speeds in the training stage (the left) and trade-offs between inference time and performance (the right) of existing mainstream methods. Our proposed one-stage method ED-Pose shows the superiority of efficiency compared with the Bottom-Up (BU) model HigherHRNet~\citep{cheng2020higherhrnet}, Top-Down (TD) models Sim.Base.~\citep{xiao2018simple} and DETR-based Poseur~\citep{mao2022poseur}, the one-stage method PETR~\citep{shi2022end}.
%Besides HrHRNet, all methods use the same backbone ResNet50.
}
\label{fig:convergence} 
% \vspace{-0.1cm}
\end{figure}

For the inference time, ED-Pose surpasses all the bottom-up and one-stage methods in both speed and accuracy fields from Table.~\ref{tab:sota} and ~\ref{tab:one-stage}. 
Notably, from the right of Fig.~\ref{fig:convergence}, we compare the inference time under different AP scores. ED-Pose achieves the best efficiency and effectiveness trade-offs.
Under the competitive performance (about $74.2$) with the DETR-Based top-down method (Poseur with RLE loss~\citep{li2021human}), ED-Pose can boost inference speed by $30.7\%$. 
%Although our ED-Pose is behind some top-down methods in terms of accuracy (e.g. Poseur with RLE loss~\citep{li2021human}), ED-Pose with ResNet-50 has great superiority in inference efficiency (by . 
%
Moreover, the left of Fig.~\ref{fig:convergence} shows the convergence speeds in the training stage that are rarely discussed before. EP-Pose introducing explicit detection is faster than all methods and obtains better performance under the early epochs. Interestingly, all DETR-based methods are slower than even bottom-up methods.

% \vspace{-0.2cm}
\subsection{Ablation Study}
% \vspace{-0.2cm}
 %We also analyze keypoint detection representation, the interactive learning strategy between the human and keypoint detection, and the impact of hyperparameters on the COCO \texttt{val2017} dataset.

\textbf{Explicit human detection}: We first analyze the
effectiveness of explicit human detection on both the COCO \texttt{val2017} and the CrowdPose \texttt{test}, using the ResNet-50 backbone. We remove all human detection losses in  decoder layers to verify its effectiveness. Results in Table.~\ref{tab:box} clearly verify that explicit human detection supervision significantly improves the convergence speed of keypoint detection and precision, yielding +$4.5$ AP on COCO and +$9.9$ AP on CrowdPose.

\begin{table*}[h]
\vspace{-0.7cm}
\centering
\caption{Impact on explicit human detection (Human Det.) for convergence speed and precision on the COCO dataset (the left) and CrowdPose dataset (the right).}
    \begin{center}
        \begin{minipage}[t]{0.46\textwidth}
        \centering
        \resizebox{\linewidth}{!}{
            \makeatletter\def\@captype{table}\makeatother
                \begin{tabular}{c|cccccc}
                    \toprule
                       {\textbf{Human Det.}} & 12e & 24e & 36e & 48e & 60e \\
                    \midrule
                     & 41.1 & 56.6& 61.0 & 64.7 & 67.1    \\
                     $\checkmark $ & \textbf{60.5}&  \textbf{67.5} &  \textbf{69.7}  & \textbf{70.8} & \textbf{71.6} \\
                  \bottomrule
                \end{tabular}
            }
            \label{tab:hw}
        \end{minipage}
        \hspace{-0.5cm}
        \quad
        \begin{minipage}[t]{0.51\textwidth}
        \centering
        \resizebox{\linewidth}{!}{
            \makeatletter\def\@captype{table}\makeatother
                    \begin{tabular}{c|cccccc}
                        \toprule
                           {\textbf{Human Det.}} & 12e & 24e & 36e & 48e & 60e & 80e\\
                        \midrule
                         & 13.1 & 20.5& 31.1 & 42.6 & 51.3 & 60.0   \\
                         $\checkmark $ & \textbf{37.1} & \textbf{54.8} &  \textbf{62.4}  & \textbf{66.4} & \textbf{68.2} & \textbf{69.9} \\
                      \bottomrule
                    \end{tabular}
            }
            \label{tab:mask}
        \end{minipage}
    \end{center}
     \label{tab:box}
\vspace{-0.3cm}
\end{table*}

% the performance drop caused by removing the above loss shows that explicit human detection supervise can benefit the initialization of the latter keypoint detection and the interaction between human and keypoint detection.

% ED-Pose introduces the explicit human detection via the human detection decoder, which influence a good initialization for the latter keypoint detection, as well as the interactive learning between human and keypoint detection. Thus, we remove the human detection related loss in all the human detection decoder layers and human-to-keypoint detection decoder layers to verify the effectiveness of explicit human detection as shown in Table.~\ref{tab:box}.

\textbf{Keypoint detection representation}: 
% To explore the effect of the initialization of the defined 4D keypoint detection representation $(x,y,w,h)$, we conduct five ablation under different initialization ways:1) 
ED-Pose reformulates pose estimation as the keypoint box detection, where the 2D center-point coordinate $(x,y)$ of keypoint is extended to the 4D representation $(x,y,w,h)$. As the region of feature aggregation for keypoint depends on the width and height of the keypoint box, we explore the effect of $(w,h)$  under five initialization ways: 1). \textbf{None} discards width and height and keeps the original 2D coordinate representation
%to select its corresponding features via the deformable attention
; 2). \textbf{Min.} denotes that the keypoint is initialized by $1\%$ width and height of the corresponding human box; 3). \textbf{Max.} means that a keypoint is directly initialized by the width and height of its corresponding human box; 4). \textbf{FFN.} is that we apply an FFN network upon the human content query to regress a 2D weight of each keypoint within the human box
%, which are used to multiply the width and height of the human box to get the keypoint box
; 5). \textbf{Ours} means that we initialize learnable embeddings for each keypoint across the whole dataset and utilize it to weigh the width and height of the human box. 
As shown in Table.~\ref{tab:hw}, the result without width and height representation is much lower than those of the other four settings, leading to a $13.3$ AP drop compared with the best initialization way (Ours) since it will lose contextual information to query the local and global relations. Learnable width and height obtain better performance than the fixed sizes. 

\textbf{The interactive learning between human and keypoint detection}:
Our human-to-keypoint detection decoder enhances the interactivity among external human-to-human, internal human-to-keypoint, and internal keypoint-to-keypoint. Thus, we explore the impact of different interactive learning strategies: 1) \textbf{Full} makes all human queries and keypoint queries interact one by one; 2) \textbf{w/o H-K} removes the internal human-to-keypoint interaction; 3) \textbf{w/o H-H} discards the external human-to-human interaction; 4) \textbf{Ours} is our full interactive learning strategy. As shown in Table.~\ref{tab:mask}, direct full connection for all queries could lead to $1.2$ AP degradation. The reason is that external keypoint-keypoint interaction in a full connection strategy may cause a large disturbance, especially in crowded situations. Besides, other interactive strategies slightly affect performance as well.

\textbf{Hyperparameter tuning}: To explore the effect on the number $M$ of selected queries in Sec.~\ref{sec:N-to-M}, we conduct three numbers with $50$, $100$, and $200$, respectively. As shown in Table.~\ref{tab:query}, increasing selected human queries can improve performance, but it meets a bottleneck for further improvement.

\begin{table*}[h]
\vspace{-0.6cm}
    \begin{center}
        \begin{minipage}[t]{0.37\textwidth}
        \centering
        \caption{Impact on the width and weight initialization of the keypoint box.}
        \resizebox{\linewidth}{!}{
            \makeatletter\def\@captype{table}\makeatother
              \begin{tabular}{l|ccccc}
                \toprule
                   {{$\mathbf{(w,h)}$}} & None & Min. & Max. & FFN. & Ours \\
                \midrule
                  ${\rm AP}$ & 58.3  & 70.8 &  70.8 & 71.2 & \textbf{71.6} \\
                  ${\rm AP}_{M}$ & 57.5 & 65.1 &  64.9 & 65.7 & \textbf{65.9} \\
                  ${\rm AP}_{L}$ & 60.0 &  79.2 & 79.3 & 79.2 & \textbf{79.8}\\
              \bottomrule
              \end{tabular}
            }
            \label{tab:hw}
        \end{minipage}
        \hspace{-0.25cm}
        \quad
        \begin{minipage}[t]{0.37\textwidth}
        \centering
        \caption{Impact on interactive learning between human and keypoint detection.}
        \resizebox{\linewidth}{!}{
            \makeatletter\def\@captype{table}\makeatother
              \begin{tabular}{l|ccccc}
                \toprule
                   {\textbf{Strategy}} & Full & w/o H-K & w/o H-H & Ours \\
                \midrule
                ${\rm AP}$ & 70.4  & 71.2 &  71.3 & \textbf{71.6} \\
                  ${\rm AP}_{M}$ & 64.6 & 65.6& \textbf{66.0} & 65.9\\
                  ${\rm AP}_{L}$ & 78.8 & 79.3 & 79.0 & \textbf{79.8} \\
              \bottomrule
            \end{tabular}
            }
            \label{tab:mask}
        \end{minipage}
        \hspace{-0.25cm}
        \quad
        \begin{minipage}[t]{0.22\textwidth}
        \centering
        \caption{Impact on the number of $M$ selected queries.}
        \resizebox{\linewidth}{!}{
            \makeatletter\def\@captype{table}\makeatother
            \begin{tabular}{l|ccc}
            \toprule
               {\textbf{$M$}} & $50$ & $100$  & $200$\\
            \midrule
                ${\rm AP}$ & 70.9 & \textbf{71.6} & \textbf{71.6}  \\
                  ${\rm AP}_{M}$ & 65.2 & 65.9 & \textbf{66.0} \\
                  ${\rm AP}_{L}$ & 79.2 & \textbf{79.8} & 79.6 \\
            \bottomrule
            \end{tabular}
            }
            \label{tab:query}
        \end{minipage}
    \end{center}
\vspace{-0.4cm}
\end{table*}

\section{Conclusion}
\vspace{-0.1cm}
In this work, we re-consider the multi-person pose estimation task as two explicit box detection processes. We unify the global person and local keypoint into the same box representation, and they can be optimized by the consistent regression loss in a fully end-to-end manner. Based on the novel methods, the proposed ED-Pose surpasses existing end-to-end methods by a large margin and shows superiority over the long-standing two-stage methods on both COCO and CrowdPose.
This is a simple attempt to make the whole pipeline succinct and effective. We hope this work could inspire further one-stage designs.

%limitation: bbox AP虽然比TD用到的高，但最后kpt AP依然不如TD， 说明这里还有很大进步空间
\vspace{-0.2cm}
\section*{Acknowledgement}
\vspace{-0.2cm}
The work is partially supported in by the Young Scientists Fund of the National Natural Science Foundation of China under grant No. 62106154, by the Natural Science Foundation of Guangdong Province, China (General Program) under grant No.2022A1515011524, and by Shenzhen Science and Technology Program ZDSYS20211021111415025.

\bibliographystyle{iclr2023_conference}
\bibliography{iclr2023_conference}

\appendix

\newpage

\begin{center}
		\textbf{\Large Appendix:\\Explicit Box Detection Unifies End-to-End Multi-Person Pose Estimation}\end{center}

In this Appendix, we provide descriptions of a detailed experimental setup (in Sec.~\ref{sec:app_exp}), more comparisons under COCO test-dev (in Sec.~\ref{sec:app_cocotest}), results of human detection on both COCO and CrowdPose datasets (in Sec.~\ref{sec:app_det}), more visualization to show the effect of explicit box detection and comparison with PETR (in Sec.~\ref{sec:app_viz}), \Revise{more discussion for ED-Pose (in Sec.~\ref{sec:app_dis})}. We also append our code to reproduce the results.

\appendix

\section{Experiment Setup}
\label{sec:app_exp}
\textbf{Dataset.}
Our experiments are mainly conducted on the popular COCO2017 Keypoint Detection benchmark~\citep{lin2014microsoft}, which contains about $250K$ person instances with $17$ keypoints. We compare with other state-of-the-art methods on both the \texttt{val2017} set and \texttt{test-dev} set. 
To verify the superiority of explicit detection, we also evaluate our approach on the CrowdedPose dataset~\citep{li2019crowdpose} which is more challenging and includes many crowded and occlusion scenes. It consists of $20K$ images containing about $80K$ persons with $14$ keypoints. 
%We train our models on the \texttt{train} and \texttt{val} sets and report the results on the \texttt{test} set. 
For ablation studies, we report results on the COCO \texttt{val2017} set.
The OKS-based Average Precision (AP) is employed as the main evaluation metric on both datasets.

\textbf{Implementation details.} In the training stage, we augment input images by random crop, random flip, and random resize with the shorter sides in $[480,800]$ and the longer sides less or equal to $1333$ following DETR~\citep{carion2020end} and PETR~\citep{shi2022end}. To accelerate the early explicit human detection, we use a human query denoising training strategy from DN-DETR~\citep{li2022dn}.
We use the AdamW~\citep{kingma2014adam,loshchilov2017decoupled} optimizer with weight decay of $1 \times 10^{-4}$ and train our model on Nvidia A100 GPUs with batch size 16 for $60$ epochs and $80$ epochs on COCO and CrowdPose, respectively. The initial learning rate is $1 \times 10^{-4}$ and is decayed at the $55$th epoch and $75$th epoch by a factor of 0.1 on COCO and CrowdPose, respectively. The channel dimension $D$ is set to 256. \Revise{The number of layers in Human Detection Decoder and Human-to-Keypoint Detection Decoder are $2$ and $4$ respectively.} 
In the test stage, the input images are resized to have their shorter sides being $800$ and their longer sides less or equal to $1333$. 

\Revise{\textbf{Loss Function.} 
the overall loss function of ED-Pose can be formulated as:
\begin{gather}
        L=L_h+L_c+L_k \\
        L_h=\mu \left|H-H^*\right|+\beta(1-\mathrm{GIOU}) \\
        L_c=-\lambda \alpha(1-p_t)^{\gamma}log(p_t), \mathrm{where}~p_t=p~\mathrm{if}~y=1,~p_t=1-p~\mathrm{if}~y\neq1  \\
        L_k= \omega \left|P-P^*\right| + \theta \frac{\sum^{K}_i\mathrm{exp}(-\left|P_i-P_i^*\right|/2s^2k_i^2)\delta(v_i>0)}{\sum^{K}_i\delta(v_i>0)} 
\end{gather}}
\Revise{where $L_h$ is for human box regression that contains L1 loss and GIOU~\citep{rezatofighi2019generalized} loss, $L_c$ is for human classification that is focal loss~\citep{lin2017focal} with $\alpha$ = 0.25, $\gamma$ = 2, and $L_k$ is for keypoint regression that includes L1 loss and the constrained L1 loss-OKS loss~\citep{shi2022end}. $\left|H-H^*\right|$ is the L1 distance between the predicted human boxes and the ground-truth ones. $y\in{\pm1}$ specifies the ground-truth class, and $p\in[0, 1]$ is the ED-Pose's estimated probability for the class with label $y=1$. $\left|P-P^*\right|$ is the L1 distance between predicted keypoints inside a human and the ground-truth ones. $\left|P_i-P_i^*\right|$ is the L1 distance between the $i$-th predicted keypoint and ground-truth one, $v_i$ is the visibility flag of the ground truth, $s$ is the object scale, and $k_i$ is a per-keypoint constant that controls falloff.}

\Revise{The loss coefficients $\mu, \beta, \lambda, \omega, \theta$ are $5$, $2$, $2$, $10$, $4$.}
% We employ the L1 loss and GIOU loss \citep{rezatofighi2019generalized} for human box regression, focal loss \citep{lin2017focal} with $\alpha=0.25$, $\gamma=2$ for human classification, L1 loss and the constrained L1 loss named OKS loss \citep{shi2022end} for keypoint regression. The loss coefficients are 5.0 for L1 loss in human box regression, 2.0 for GIOU loss in human box regression, 1.0 for focal loss in human classification, 10.0 for L1 loss in keypoint regression, and 4.0 for OKS loss in keypoint regression.

\textbf{The Detailed Interactive Learning between Human Detection and Keypoint Detection.}
The relations from internal human-keypoint, internal keypoint-keypoint, and external human-human are learned from the self-attention mechanism. 
As shown in Figure.~\ref{fig:decoder}-(c), Human-to-Keypoint Interactive Attention can be computed as follow:
% \begin{gather}
% \mathbf{Q}_H^c(i)'=\frac{f(\mathbf{Q}_H^c(i)+\mathrm{PE}(\mathbf{Q}_H^p(i)))\cdot f(\mathbf{Q}_K^c(i)+\mathrm{PE}(\mathbf{Q}_K^p(i)))^\mathrm{T}}{\sqrt{D}}\cdot f(\mathbf{Q}_H^c(i))\\
% \mathbf{Q}_K^c(i)'=\frac{f(\mathbf{Q}_K^c(i)+\mathrm{PE}(\mathbf{Q}_K^p(i)))\cdot
% f(\mathbf{Q}_H^c(i)+\mathrm{PE}(\mathbf{Q}_H^p(i)))^\mathrm{T}}{\sqrt{D}}\cdot f(\mathbf{Q}_K^c(i))
% \end{gather}
\begin{gather}
    \mathbf{Q}_{H,K}^{c'}=\mathrm{softmax}(\frac{f(\mathbf{Q}_{H,K}^c+\mathrm{PE}(\mathbf{Q}_{H,K}^p))\cdot f(\mathbf{Q}_{H,K}^c+\mathrm{PE}(\mathbf{Q}_{H,K}^p))^\mathrm{T}}{\sqrt{D}}+\mathcal{M})\cdot f(\mathbf{Q}_{H,K}^c)
\end{gather}
where $\rm{PE}$ denotes positional encoding, $f(\cdot)$ is linear projection, $\mathbf{Q}_{H,K}^c \in \mathbb{R}^{(M+M*K)\times D}$ and $\mathbf{Q}_{H,K}^p \in \mathbb{R}^{(M+M*K)\times4}$ are human-keypoint content queries and human-keypoint position queries for $M$ human candidates and the corresponding $K$ keypoints. $D$ is 256 by default. In practice, to implement the three interactive learning processes in a simple way, we use an attention mask $\mathcal{M} \in \mathbb{R}^{(M+M*K)\times(M+M*K)}$ to block the interactiveness between external keypoint-keypoint. Thus, $\mathcal{M}$ can be formulated as:
\begin{equation}
\mathcal{M}(i,j)=
\begin{cases}
0& \mathcal{M}(i,j)=\mathrm{True} \\
-\infty& \mathcal{M}(i,j)=\mathrm{False}
\end{cases}
\end{equation}
where $\mathcal{M}(i,j)$ is the location of the attention mask. We use the $\mathcal{M}$ to keep internal human-keypoint attention (True), internal keypoint-keypoint attention (True), and external human-human attention (True) while avoiding external keypoint-keypoint attention (False). 

\Revise{\textbf{Inference time.} 1) Comparison Methods: All of the methods in Table.~\ref{tab:sota} are tested on our A100 machine for a fair comparison and the detector of top-down methods is tested by MMdetection. In tabel.~\ref{tab:one-stage}, the inference time with \emph{\textbf{emphasized bold number}} is tested by our A100 machine, otherwise, it is from PETR's paper using a V100 GPU due to no available source code.  2) Testing Rules: We omit the time for data pre-processing and only measure the time for model forwarding and data post-processing (\textit{i.e.}, grouping operation in bottom-up methods). For bottom-up methods and one-stage methods, we set the batch size to $1$. For top-down methods, we set the batch size to $5$ for simulating the multi-person situation.}

\Revise{\textbf{A Unique Merit of ED-Pose for Human Detection Pre-training.} 
Thanks to the explicit introduction of human detection, ED-Pose has a unique merit compared to other existing two-stage methods and one-stage methods, which can pre-train human detection to improve the detection performance to assist the subsequent pose estimation in a fully end-to-end manner.
Objects365 \citep{shao2019objects365} is a large-scale detection dataset with over 1.7M annotated images for training and 80, 000 annotated images for validation. 
% To use the data more efficiently, We select the first 5, 000 out of 80, 000 validation images as our validation set and add the others to training. 
We pre-train the human detection of ED-Pose (\textit{i.e.}, backbone, encoder, human detection decoder) on Objects365 for 26 epochs using 64 Nvidia A100 GPUs.}

\section{Comparison of COCO Test-dev.}
\label{sec:app_cocotest}
From Table.~\ref{tab:one-stage}, our method without dense heatmap loss can significantly outperform all existing one-stage methods, such as DirecetPose~\citep{tian2019directpose}, FCPose~\citep{WeianMao2021FCPoseFC}, Inspose~\citep{DahuShi2021InsPoseIN}, CenterNet~\citep{zhou2019objects}, and PETR~\citep{shi2022end}. To be specific, our proposed ED-Pose has $2.2$ AP gains compared with the fully end-to-end method PETR~\citep{shi2022end} with a ResNet-50 backbone or Swin-L backbone with about $50\%$ inference time reduction.

\label{sec:app_coco}
\begin{table*}[h]
\centering
\vspace{-0.4cm}
\caption{Comparisons with state-of-the-art one-stage methods on COCO \texttt{test-dev} dataset. \underline{underlined} highlights the compared results. \Revise{\emph{Emphasized bold number} in inference time is tested by our A100 machine}.}
\resizebox{0.9\linewidth}{!}{
    \begin{threeparttable}   
  \begin{tabular}{p{0.4cm}|l|l|l|ccc|cc|c}
    \toprule
       \multicolumn{2}{c|}{\textbf{Method}} & Backbone & Loss & ${\rm AP}$ & ${\rm AP}_{50}$ & ${\rm AP}_{75}$ & ${\rm AP}_{M}$ & ${\rm AP}_{L}$ & Time [ms]\\
    %   \midrule
    %   \multicolumn{10}{c}{\textbf{One-stage Methods}} \\
    \midrule
    \multirow{7}*{{\small{\rotatebox{90}{Non E-2-E}}}}
    &DirectPose & ResNet-50 & R.& 62.2 & 86.4  & 68.2  & 56.7 & 69.8 & 74\ \\
    &DirectPose & ResNet-101 & R.& 63.3 & 86.7 &  69.4 & 57.8 & 71.2 & -\ \\
    &FCPose  & ResNet-50 & R+HM.   & 64.3 & 87.3 & 71.0 & 61.6 & 70.5 & 68 \\
    & FCPose & ResNet-101 & R+HM.  & 65.6 & 87.9 & 72.6 & 62.1 & 72.3 & 93\\
    & InsPose & ResNet-50 &  R+HM.  & 65.4 & 88.9 & 71.7 & 60.2 & 72.7 & \Revise{\emph{\textbf{78}}}\\
    &InsPose & ResNet-101 &R+HM.  &  66.3 & 89.2 & 73.0 & 61.2 & 73.9 & 100\\
    &CenterNet & Hourglass & R.& 63.0 & 86.8 & 69.6 & 58.9 & 70.4 & 160\\
        \midrule
    \multirow{4}*{{\small{\rotatebox{90}{Fully E-2-E}}}}
    &PETR & ResNet-50 &R+HM.  &  \underline{67.6} & 89.8 & 75.3 & 61.6 & 76.0 & \emph{\textbf{105}}\\
    &PETR & Swin-L & R+HM.  & \underline{70.5} & 91.5 & 78.7 & 65.2 & 78.0 & \emph{\textbf{206}}\\
    \cmidrule{2-10}
    &ED-Pose & ResNet-50 & R.  &  69.8{\color{Red}$\uparrow_{2.2}$} & 90.2 & 77.2 & 64.3 & 77.4& \emph{\textbf{{51}}}{\color{Red}$\downarrow_{51.4\%}$}\\
    &ED-Pose & Swin-L & R.  &  \textbf{72.7}{\color{Red}$\uparrow_{2.2}$} & \textbf{92.3} & \textbf{80.9} & \textbf{67.6} & \textbf{80.0}& \emph{\textbf{88}}{\color{Red}$\downarrow_{57.3\%}$} \\
   % &ED-Pose & Swin-L* & R.  &  & & & & & \underline{142}{\color{Red}$\downarrow_{31.1\%}$}\ 
  \bottomrule
\end{tabular}
\begin{tablenotes}
\item[1] \Revise{The inference time without \emph{Emphasized bold number} is from PETR's paper because there is no public code for replication.}
\end{tablenotes}
  \end{threeparttable}}
    \label{tab:one-stage}
\end{table*}

\section{Results of Human Detection}
\label{sec:app_det}
Due to the explicit person detection in our methods, we also report the human detection performance on the COCO \texttt{val2017} set and CrowdPose \texttt{test} set in Figure.~\ref{tab:box2}. 
For the COCO dataset, ED-Pose with Swin-L* backbone gains $7.7$ ${\rm AP}_{M}$ and $6.7$ ${\rm AP}_{L}$ improvement compared with Faster-RCNN~\citep{ShaoqingRen2015FasterRT}. For the CrowdPose dataset,
most top-down methods directly use YOLOv3~\citep{redmon2018yolov3} pre-trained on COCO \texttt{trainval} to generate the detected human bounding boxes, 
making their performance even worse than recent bottom-up methods.
ED-Pose can provide human detection results with nearly doubling performance on ${\rm AP}_{M}$ and ${\rm AP}_{L}$ to serve future work. 
Moreover, in Table~\ref{tab:layer}, we study the impact of different decoders (\textit{i.e.}, human detection decoder and human-keypoint detection decoder) for human detection and keypoint detection. 
First, we discover that the keypoint initialization provided by the human detection decoder has already achieved good results on COCO and CrowdPose datasets, which proves the effectiveness of explicit human detection. 
Second, the keypoint detection introduced via the human-keypoint detection decoder is also helpful to improve large-scale human detection and obtain similar performance on medium-scale human detection. 
Different from previous observations of severe  optimization conflicts between human detection and keypoint detection, ED-Pose unifies the contextual learning between human-level and keypoint-level information and gains benefits from each other. 
%

% Notably, since small objects in the COCO dataset and CrowdPose dataset are not labeled with keypoints, we only compare ${\rm AP}_{M}$ (medium object) and ${\rm AP}_{L}$ (large object) here. 
% Compared with the commonly used human detection methods, \textit{i.e.}, Faster-RCNN~\citep{ShaoqingRen2015FasterRT} and YOLOv3~\citep{redmon2018yolov3}, ED-Pose can provide 

\begin{table*}[h]
\vspace{-0.4cm}
\caption{The Average Precision comparisons of the commonly used human detection methods (e.g., Faster-RCNN and YOLOv3) with ours on COCO (the left) and CrowdPose (the right). Notably, We only compare ${\rm AP}_{M}$ (medium object) and ${\rm AP}_{L}$ (large object) here as small objects in the two datasets are not labeled with keypoints. ED-Pose can provide better human detection to serve future work, especially for top-down and one-stage methods}
    \begin{center}
        \begin{minipage}[t]{0.35\textwidth}
        \centering
        \resizebox{\linewidth}{!}{
            \makeatletter\def\@captype{table}\makeatother
                \begin{tabular}{l|cc}
                    \toprule
                       {\textbf{Methods}}  &${\rm AP}_{M}$ & ${\rm AP}_{L}$  \\
                    \midrule
                        Faster-RCNN &  63.3 & 74.5  \\
                      ED-Pose (ResNet-50) &  65.9 & 77.6  \\
                         ED-Pose (Swin-L)  &69.7 & 80.2 \\
                         ED-Pose (Swin-L*)  &\textbf{71.0} & \textbf{81.2} \\
                  \bottomrule
                \end{tabular}
            }
            %\label{tab:hw}
        \end{minipage}
        \hspace{-0.5cm}
        \quad
        \begin{minipage}[t]{0.35\textwidth}
        \centering
        \resizebox{\linewidth}{!}{
            \makeatletter\def\@captype{table}\makeatother
                    \begin{tabular}{c|cccccc}
                        \toprule
                        %    {\textbf{Methods}} & ${\rm AP}$ & ${\rm AP}_{50}$ & ${\rm AP}_{75}$ & ${\rm AP}_{S}$ & ${\rm AP}_{M}$ & ${\rm AP}_{L}$  \\
                        % \midrule
                        %  YOLOv3 & 39.6 & 69.0 & 42.5 & \textbf{20.9} & 37.4 & 46.5   \\
                        %   ED-Pose (ResNet-50) & 60.0 & 78.6 & 65.9 & 12.6 & 58.0 & 75.3   \\
                        %  ED-Pose (Swin-L)  &63.2 & 80.4 &  69.5  & 14.2 & 61.8 & 78.4 \\
                        %  ED-Pose (Swin-L*)  &\textbf{66.6} & \textbf{81.3} &  \textbf{72.8}  & 15.1 & \textbf{66.8} & \textbf{82.2} \\
                                                   {\textbf{Methods}} & ${\rm AP}_{M}$ & ${\rm AP}_{L}$  \\
                        \midrule
                         YOLOv3  & 37.4 & 46.5   \\
                          ED-Pose (ResNet-50)  & 58.0 & 75.3   \\
                         ED-Pose (Swin-L)   & 61.8 & 78.4 \\
                         ED-Pose (Swin-L*)  & \textbf{66.8} & \textbf{82.2} \\
                      \bottomrule
                    \end{tabular}
            }
            % \label{tab:mask}
        \end{minipage}
    \end{center}
     \label{tab:box2}
\vspace{-0.4cm}
\end{table*}

\begin{table*}[h]
\vspace{-0.4cm}
\centering
\caption{The performance comparisons of results from different decoders (\textit{i.e.}, human detection decoder and human-keypoint detection decoder) on both COCO and CrowdPose datasets with the Swin-L as the backbone. }
\resizebox{1\linewidth}{!}{
  \begin{tabular}{l|l|l|cc|ccc}
    \toprule
       \textbf{Method} & Dataset & Decoder &  ${\rm AP}_{M}$ (Human) & ${\rm AP}_{L}$ (Human) & ${\rm AP}$ (Keypoint)  \\
       \midrule
           ED-Pose & COCO & Human Det. &  \textbf{70.4}  & 79.6 & 70.4  \\
          ED-Pose & COCO & Human Det.+Human-Keypoint Det. & 69.8  & \textbf{80.1} & \textbf{74.3}   \\       \midrule
                     ED-Pose & CrowdPose & Human Det. &  \textbf{61.7}  & 76.9 & 68.0    \\
          ED-Pose & CrowdPose & Human Det.+Human-Keypoint Det. & 61.6  & \textbf{78.4} & \textbf{73.1}    \\ 
       \midrule
\end{tabular}}
\label{tab:layer}
\vspace{-0.4cm}
\end{table*}

\section{Qualitative Results}
\label{sec:app_viz}
\textbf{Qualitative ablations for explicit human detection}:
In the main article, we verify the effectiveness of explicit human detection for convergence speed and precision of the keypoint detection. Here, we present the corresponding qualitative results on the CrowdPose \texttt{test} set as shown in Fig.~\ref{fig:viz2}. In general, explicit human detection has several advantages: 1). It can provide global human-level information, enabling the model to be aware of flipping easily like the case in the first row. 2). It can give a clear human identity to the crowded scene to avoid the keypoint mismatching across persons, such as the case in the second row. 3) It is friendly for small human pose estimation as it can provide a precise prior of human box position, e.g., the case in the third row. 

\textbf{Qualitative comparisons between PETR and ED-Pose}:
We present the visualization comparisons of ED-Pose and PETR~\citep{shi2022end} on the COCO dataset, as shown in Fig~\ref{fig:viz3}. From the first row, ED-Pose can pay attention to the flipping issue due to the introduction of explicit detection and the realization of human-keypoint feature propagation. The second row further reflects that explicit detection in ED-Pose can make the model aware of the human position when conducting pose estimation, rather than the chaotic and unconscious query on a full image like PETR. All possible detected keypoints from PETR have low scores, making it hard to distinguish the right and wrong results. The third row shows that the enhancement of the global-local interactivity could relieve the estimation problem of the hard pose with heavy occlusions.

\section{More Discussion}
\label{sec:app_dis}

\Revise{\textbf{Discussion for Additional Related Works.} (1) N. xue \citep{xue2022learning} focuses on improving the performance of the heatmap-based bottom-up method by learning the local-global contextual adaptation without any explicit box detection. (2) Different from the one-stage method, D. Wang \citep{wang2022contextual} decouples persons into multiple instance-aware feature maps to obtain the keypoints without any explicit box detection, and they also use heatmap-based supervision.}

\Revise{Different from their works, ED-Pose aims to utilize two explicit box detection processes with a unified box representation and light L1 losses. It abandons all post-processings in a fully end-to-end manner. In terms of performance, ED-Pose surpasses them by a margin and even outperforms heatmap-based Top-down methods under the same backbone.}

\begin{figure}[h]	
\vspace{-0.4cm}
\centering
 	{
 		\begin{minipage}[t]{1\linewidth}
 			\centering         
 			\includegraphics[width=1\linewidth]{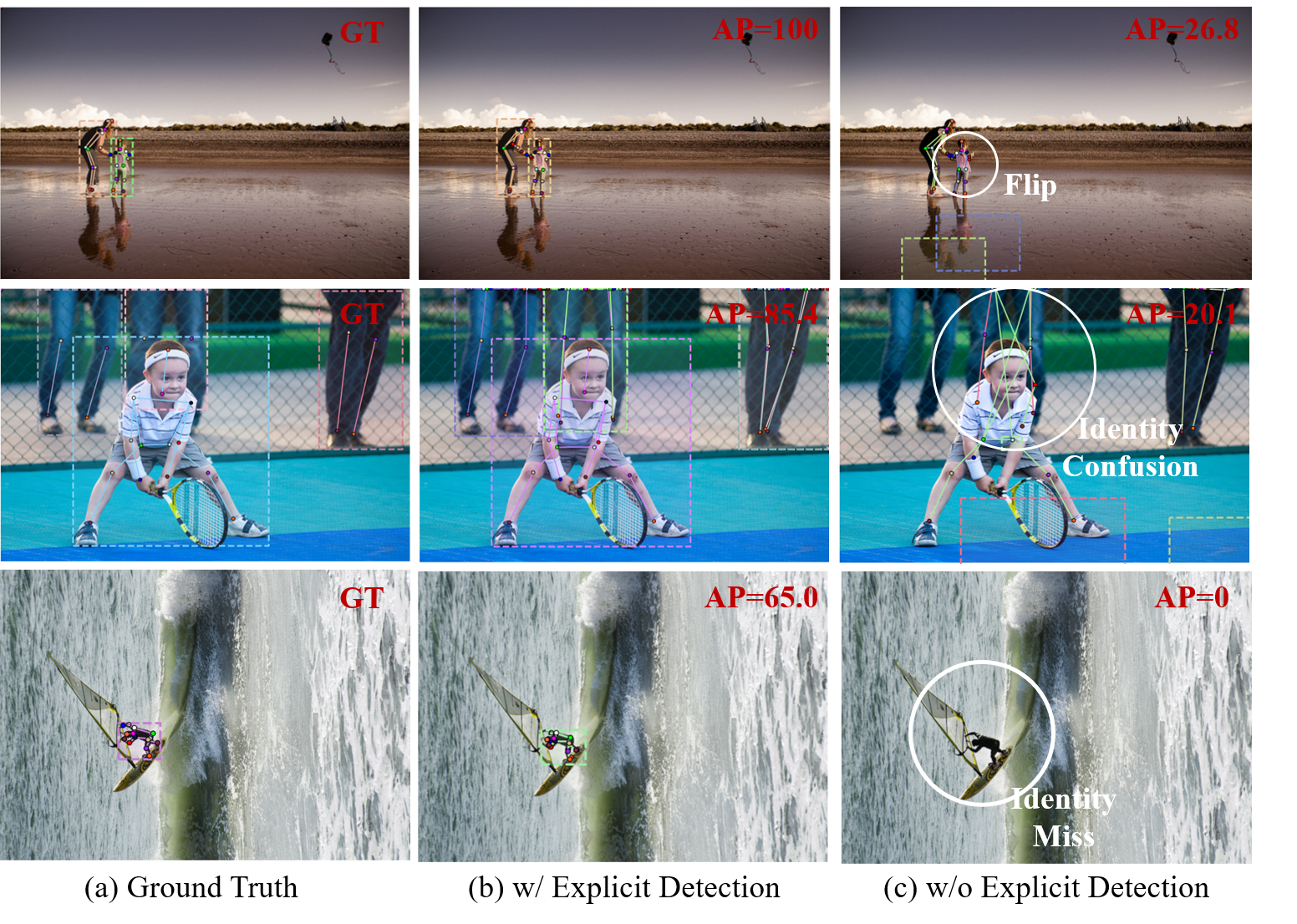}  
 		\end{minipage}
 	} 
\vspace{-0.6cm}
\caption{Visualization results of the ablation study for explicit human detection on the CrowdPose dataset. Notably, we ignore the keypoint box to compare the keypoint localization of the two settings clearly. The white circle is the failure part with a description of the type of failure.}
\label{fig:viz2} 
\vspace{-0.3cm}
\end{figure}

\begin{figure}[h]	
\vspace{-0.2cm}
\centering
 	{
 		\begin{minipage}[t]{1\linewidth}
 			\centering         
 			\includegraphics[width=1\linewidth]{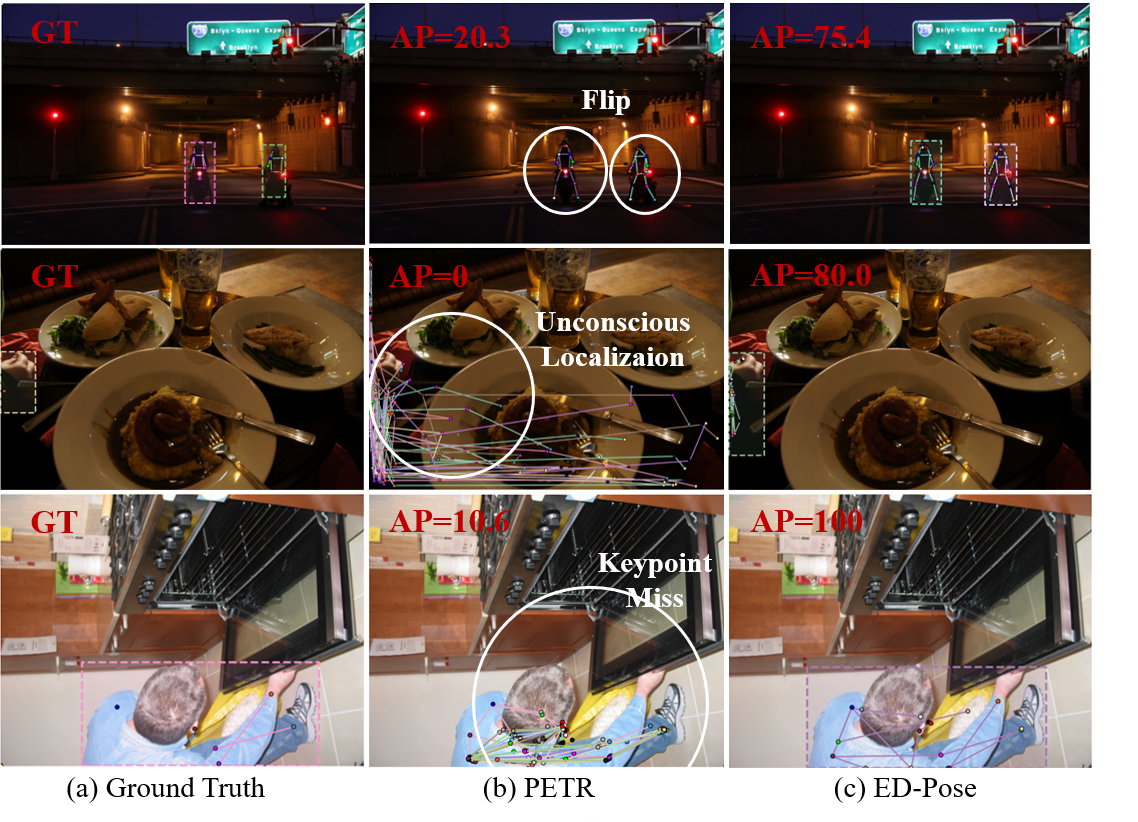}  
 		\end{minipage}
 	} 
\vspace{-0.6cm}
\caption{Visualization results of PETR and ED-Pose on COCO dataset. Both methods are based on ResNet-50 as the backbone. Notably, there are several candidates in the PETR's result at the second line as their classification scores are close and relatively low.}
\label{fig:viz3} 
% \vspace{-0.6cm}
\end{figure}

\Revise{\textbf{The More Detailed Advantage of ED-Pose.}
From the results, ED-Pose is efficient and effective even with the same backbone compared with previous works, indicating it really helps with pose estimation.
From the method with two box detection processes, we have admitted it is inspired by recent DETR-based methods. However, the motivation and usage have obvious differences due to the task differences between object detection and multi-person pose estimation. Multi-person pose estimation focuses on either human-level (global) or keypoint-level (local) information. ED-Pose re-considers this task as two explicit box detection processes with a unified representation and regression supervision.
Moreover, we also compare with other DETR-based pose estimators, and ED-Pose shows its superiorities in both efficiency and performance. The first human box detection can extract global features and provide a good initialization for the latter keypoint detection, making the training process converge fast. The second keypoint box detection can bring in local contextual information near keypoints. 
However, previous DETR-based pose estimation methods only directly regressed the keypoints' 2D coordinates, we argue that the keypoint representation as a point lacks local contextual information. They directly regress local keypoints without global feature extraction, making the convergence slow and inaccurate estimation. In Figure.~\ref{fig:viz2} and Figure.~\ref{fig:viz3}, we demonstrate the qualitative comparisons of whether or not we needed to use explicit detection and PETR with ours. Lastly, we conduct comprehensive experiments on how explicit detection works in Table.~\ref{tab:box} and Table.~\ref{tab:hw}.}

\end{document}